\documentclass[10pt,journal,compsoc]{IEEEtran}

\usepackage{graphicx}
\usepackage{comment}
\usepackage{epsfig}
\usepackage{amsmath,amssymb} 
\usepackage{color}
\usepackage[table]{xcolor}
\usepackage{tikz}
\usepackage{comment}
\usepackage{multirow}
\usepackage[hang,flushmargin]{footmisc} 
\usepackage{booktabs,tabularx}
\usepackage{enumitem}
\usepackage{makecell}
\usepackage{algorithm}
\usepackage{algorithmic}
\usepackage{threeparttable}
\usepackage{lipsum}
\usepackage[pagebackref=false, breaklinks=true, colorlinks, bookmarks=false]{hyperref}
\usepackage{url}

\usepackage{graphicx}
\usepackage{amsmath}
\usepackage{amssymb}
\usepackage{booktabs}
\usepackage{multirow}
\usepackage{stfloats}
\usepackage{caption}
\usepackage{colortbl}
\usepackage{multirow}
\usepackage{multicol}
 \usepackage{subcaption} 
\usepackage{marvosym}
\usepackage{pifont}
\usepackage{makecell}
\usepackage{booktabs}
\usepackage{footnote}
\makesavenoteenv{table}

\definecolor{linecolor}{rgb}{0.82, 0.94, 0.75}

\newcommand{\dkliang}[1]{{\color{black} #1}}

\definecolor{lowred}{RGB}{238,18,137}

\definecolor{lowerred}{RGB}{255,110,180}

\newcommand{\dplus}[1]{\fontsize{6pt}{0.1em}\selectfont (\textbf{\textcolor{lowred}{#1}})}

\ifCLASSOPTIONcompsoc
  \usepackage[nocompress]{cite}
\else
  \usepackage{cite}
\fi

\ifCLASSINFOpdf
\else
\fi
\begin{document}

\title{SOOD++: Leveraging Unlabeled Data to Boost Oriented Object Detection}

\author{
Dingkang Liang, \textit{Graduate Student Member, IEEE}, Wei Hua, Chunsheng Shi, Zhikang Zou, \\Xiaoqing Ye, Xiang Bai, \textit{Senior Member, IEEE}
\IEEEcompsocitemizethanks{
\IEEEcompsocthanksitem Dingkang Liang and Chunsheng Shi are with the School of Artificial Intelligence and Automation, Huazhong University of Science and Technology. (dkliang, csshi)@hust.edu.cn

\IEEEcompsocthanksitem Wei Hua is with the School of Electronic Information and Communications, Huazhong University of Science and Technology. Perdredes@outlook.com

\IEEEcompsocthanksitem Zhikang Zou and Xiaoqing Ye are with Baidu Inc., China.

\IEEEcompsocthanksitem Xiang Bai is with the School of Software Engineering, Huazhong University of Science and Technology. xbai@hust.edu.cn

\IEEEcompsocthanksitem  The corresponding author is Xiang Bai (xbai@hust.edu.cn). 

}
}

\markboth{
IEEE TRANSACTIONS ON PATTERN ANALYSIS AND MACHINE INTELLIGENCE.}%
{Shell \MakeLowercase{\textit{et al.}}: Bare Advanced Demo of IEEEtran.cls for IEEE Computer Society Journals}

\IEEEtitleabstractindextext{%
\begin{abstract}

Semi-supervised object detection (SSOD), leveraging unlabeled data to boost object detectors, has become a hot topic recently. However, existing SSOD approaches mainly focus on horizontal objects, leaving oriented objects common in aerial images unexplored. At the same time, the annotation cost of oriented objects is significantly higher than that of their horizontal counterparts (an approximate 36.5\% increase in costs). Therefore, in this paper, we propose a simple yet effective Semi-supervised Oriented Object Detection method termed SOOD++. Specifically, we observe that objects from aerial images usually have arbitrary orientations, small scales, and dense distribution, which inspires the following core designs: a Simple Instance-aware Dense Sampling (SIDS) strategy is used to generate comprehensive dense pseudo-labels; the Geometry-aware Adaptive Weighting (GAW) loss dynamically modulates the importance of each pair between pseudo-label and corresponding prediction by leveraging the intricate geometric information of aerial objects; we treat aerial images as global layouts and explicitly build the many-to-many relationship between the sets of pseudo-labels and predictions via the proposed Noise-driven Global Consistency (NGC). Extensive experiments conducted on various oriented object datasets under various labeled settings demonstrate the effectiveness of our method. For example, on the DOTA-V2.0/DOTA-V1.5 benchmark, the proposed method outperforms previous state-of-the-art (SOTA) by a large margin (+2.90/2.14, +2.16/2.18, and +2.66/2.32) mAP under 10\%, 20\%, and 30\% labeled data settings, respectively, with single-scale training and testing. More importantly, it still improves upon a strong supervised baseline with 70.66 mAP, trained using the full DOTA-V1.5 train-val set, by +1.82 mAP, resulting in a 72.48 mAP, pushing the new state-of-the-art. Moreover, our method demonstrates stable generalization ability across different oriented detectors, even for multi-view oriented 3D object detectors. The project page is at \url{https://dk-liang.github.io/SOODv2/}.

\end{abstract}

}

\maketitle

\IEEEdisplaynontitleabstractindextext

%
\IEEEpeerreviewmaketitle

\ifCLASSOPTIONcompsoc
\IEEEraisesectionheading{\section{Introduction}\label{sec:introduction}}
\else
\section{Introduction}
\label{sec:introduction}
\fi

\IEEEPARstart{S}{ufficient} labeled data is essential to achieve satisfactory performance for fully-supervised object detection. However, the data labeling process is expensive and laborious. To alleviate this problem, many Semi-Supervised Object Detection~(SSOD) methods~\cite{liu2021unbiased,xu2021end,zhou2022dense,li2022pseco}, aiming to learn from labeled data as well as easy-to-obtain unlabeled data, have been proposed recently. By leveraging the potentially useful information from the unlabeled data, the SSOD methods can achieve promising improvements compared with the supervised baselines, i.e., methods only trained from the limited labeled data. 

The recent advanced SSOD methods~\cite{liu2021unbiased,xu2021end,zhou2022dense,li2022pseco} mainly focus on detecting objects with horizontal bounding boxes\footnote{Horizontal objects include both horizontally and vertically aligned ones, as they are represented by axis-aligned bounding boxes without rotation.} in general scenes. However, in more complex scenes, notably aerial scenes, we need to utilize oriented bounding boxes to precisely describe the objects. The complexity of annotating objects with oriented bounding boxes significantly surpasses that of their horizontal counterparts, e.g., the annotation process for oriented bounding boxes incurs an approximate 36.5\% increase in costs\footnote{The cost of annotating an oriented box is approximately 36.5\% more expensive than a horizontal box~(\$86 vs. \$63 per 1k)~\url{https://cloud.google.com/ai-platform/data-labeling/pricing}}. Thus, considering the higher annotation cost of oriented boxes, semi-supervised oriented object detection is worth studying.

\begin{figure*}[t]
	\begin{center}
		\includegraphics[width=0.96\linewidth]{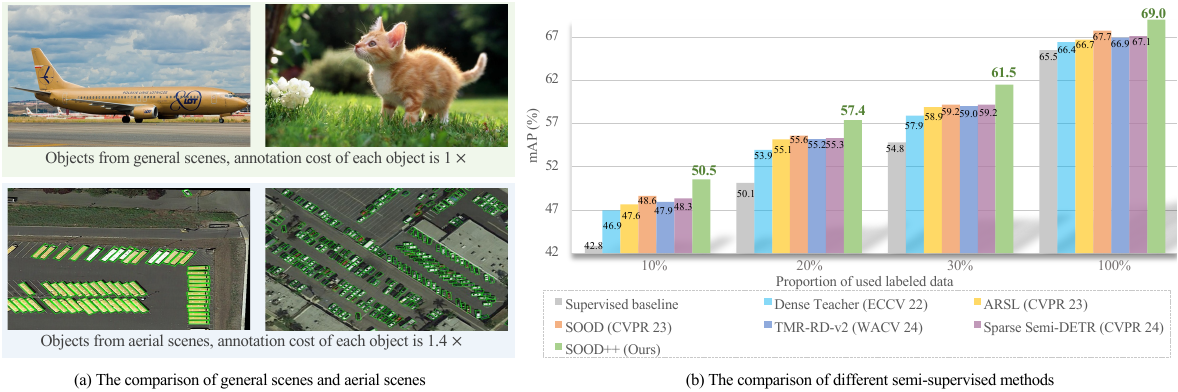}
	\end{center}
    \vspace{-10pt}
	\caption{
(a) The comparison of general scenes and aerial scenes. Different from the general objects, the objects from aerial scenes are arbitrary orientations, small, and dense distribution. Correspondingly, the annotation cost of aerial objects is higher than that of general scenes. (b) The proposed SOOD++ outperforms the supervised baseline and SOTA methods~\cite{liu2023ambiguity,shehzadi2024sparse,marvasti2024training} by a large margin on the DOTA-V1.5 benchmark (val set) under various settings.
}
	\label{fig:intro1}
\end{figure*}

Compared with general scenes, the distinguishing characteristics of objects from aerial scenes (referred to as aerial objects) are mainly from three aspects: arbitrary orientations, small, and dense distribution, as shown in Fig.~\ref{fig:intro1}(a). The mainstream SSOD methods are based on the pseudo-labeling framework~\cite{chen2022dense,tarvainen2017mean,tang2021humble,xu2021end} consisting of the teacher and student models. The teacher model operates as an Exponential Moving Average (EMA) of the student model, aggregated over historical training iterations. It aims to generate pseudo-labels for images within the unlabeled dataset, thereby facilitating the student model's learning from a combination of labeled and unlabeled data. 

However, through pilot experiments, we find that extending the SSOD framework to oriented object detection is a highly non-trivial problem. We think the following two aspects need to be considered: \textbf{1)} Geometry information is crucial for representing oriented objects, where the bounding box orientation defines the object’s direction and the aspect ratio represents the width-to-height proportion. In a semi-supervised framework, it is essential to leverage the intrinsic geometric information for effective teacher-student interaction. \textbf{2)} Aerial objects are often dense and regularly distributed in an image. These properties lead to a challenge for obtaining accurate pseudo-labels from the teacher, resulting in inaccurate knowledge transfer towards the student and damaging the learning effectiveness within a teacher-student framework. A simple yet effective solution is to consider the global perspective. The global appearance of pseudo-labels and predictions can naturally reflect their patterns and suppress local noise, benefiting a more consistent and steady learning process.

Based on these observations and analysis, in this paper, we propose a practical \textbf{S}emi-supervised \textbf{O}riented \textbf{O}bject \textbf{D}etection method termed SOOD++. SOOD++ is built upon the dense pseudo-labeling framework, where the pseudo-labels are filtered from the pixel-wise predictions~(including box coordinates and confidence scores). The key designs are from three aspects: a Simple Instance-aware Dense Sampling (SIDS) strategy, Geometry-aware Adaptive Weighting (GAW) loss, and Noise-driven Global Consistency (NGC). The SIDS is used to construct comprehensive dense pseudo-labels from the teacher, similar to pre-processing. GAW and NGC aim to measure and minimize the discrepancy between the teacher and student from an individual-level and global-level perspective, respectively.

Specifically, existing dense pseudo-labeling methods~\cite{zhou2022dense,liu2023ambiguity} directly sample the fixed number of pseudo-labels from the predicted maps, which easily overlooks the dense, small, and ambiguous objects that usually exist in aerial scenes. To construct more comprehensive pseudo-labels, we propose a Simple Instance-aware Dense Sampling (SIDS) strategy to sample the dynamic number of easy and hard cases (e.g., objects intermingled with the background) from the teacher's output. 

Considering the pseudo-label-prediction pairs are not equally informative, we propose the Geometry-aware Adaptive Weighting (GAW) loss. It utilizes the intrinsic geometry information (e.g., orientation gap and aspect ratio) of each teacher-student pair, reflecting the difficulty of the sample by weighting the corresponding loss dynamically. In this way, we can softly pick those more useful supervision signals to guide the learning of the student. 

Furthermore, we innovatively treat the outputs of the teacher and student as two independent global distributions. These distributions implicitly reflect the characteristics of the overall scenario, such as object density and spatial layout across the image. This naturally inspires us to construct global constraints, leading to the proposal of Noise-driven Global Consistency (NGC). Specifically, we first add random noise to disturb the teacher and student outputs (global distributions). Then, we use optimal transport (OT) to align the original teacher's distribution with the original student's distribution and similarly align the disturbed teacher's distribution with the disturbed student's distribution. Besides, the transport plans from the former two alignments reflect the intricate relationships within the distributions, so we propose aligning them to further evaluate noise impact and provide auxiliary guidance for the model. Applying such multi-perspective alignments will increase the models' tolerance to minor input variations, enabling them to better capture essential features of the data rather than relying excessively on noise or specific details in the unlabeled data. By promoting global consistency, our NGC not only relieves the negative effect of some inaccurate pseudo-labels in the unsupervised training stage but also encourages a holistic learning approach, considering objects' spatial relations.

Extensive experiments conducted on six real-world oriented object detection datasets in various data settings demonstrate the effectiveness of our methods. For example, as shown in Fig.~\ref{fig:intro1}(b), on the challenging large-scale DOTA-V1.5 dataset (validation set), we achieve 50.48, 57.44, and 61.51 mAP under 10\%, 20\%, and 30\% labeled data settings, significantly surpassing the state-of-the-art SSOD method~\cite{shehzadi2024sparse} by 2.14, 2.18, 2.32 mAP, respectively. More importantly, it can still improve a 70.66 mAP strong supervised baseline trained using the full DOTA-V1.5 train-val set by +1.82 mAP, resulting in 72.48 mAP on the DOTA-V1.5 dataset (test set) with single-scale training and testing. We also achieve consistent improvement on the challenging DOTA-V2.0, DOTA-V1.0, HRSC2016, DIOR-R, and even the nuScenes dataset (a representative multi-view oriented 3D object detection dataset).

In conclusion, this paper presents an early and solid exploration of semi-supervised learning for oriented object detection. By analyzing the distinct characteristics of oriented objects from general objects, we customize the key components to adapt the pseudo-labeling framework to this task. We hope this work will establish a robust foundation for the burgeoning field of semi-supervised oriented object detection and offer an effective benchmark.

\section{Related works}

\subsection{Semi-Supervised Object Detection}

In recent years, semi-supervised learning (SSL)\cite{sohn2020fixmatch,berthelot2019mixmatch} has demonstrated remarkable performance in image classification. These methods exploit unlabeled data through various techniques such as pseudo-labeling\cite{lee2013pseudo,grandvalet2004semi,xie2020self,li2023dds3d}, consistency regularization~\cite{tarvainen2017mean,xie2020unsupervised,berthelot2019mixmatch}, data augmentation~\cite{sajjadi2016regularization,chen2020simple}, and adversarial training~\cite{miyato2018virtual}. Unlike semi-supervised image classification, semi-supervised object detection (SSOD) necessitates instance-level predictions and the additional task of bounding box regression, thereby increasing its complexity. In~\cite{radosavovic2018data, zoph2020rethinking, sohn2020simple, jeong2019consistency}, pseudo-labels are generated from various augmentations.

Subsequent studies~\cite{liu2021unbiased,tang2021humble,xu2021end,yang2021interactive,liu2023mixteacher,li2022pseco} incorporate Exponential Moving Average (EMA) from Mean Teacher~\cite{tarvainen2017mean} to update the teacher model. ISMT~\cite{yang2021interactive} integrates current pseudo-labels with historical ones. Unbiased Teacher~\cite{liu2021unbiased} addresses the class-imbalance issue by replacing the cross-entropy loss with focal loss~\cite{lin2017focal}, and Unbiased Teacher v2~\cite{liu2022unbiased} uses uncertainty predictions to select pseudo-labels for the regression branch. Humble Teacher~\cite{tang2021humble} utilizes soft pseudo-labels as the student's training targets, enabling the student to distill richer information from the teacher. Soft Teacher~\cite{xu2021end} adaptively weights the loss of each pseudo-box based on classification scores and introduces box jittering to select reliable pseudo-labels. Dense Teacher~\cite{zhou2022dense} replaces post-processed instance-level pseudo-labels with dense pixel-level pseudo-labels, effectively eliminating the influence of post-processing hyperparameters. Consistent Teacher~\cite{wang2023consistent} combines adaptive anchor assignment and a Gaussian Mixture Model to reduce matching and feature inconsistencies during training. The Semi-DETR~\cite{zhang2023semi} method is specifically designed for DETR-based detectors, mitigating the training inefficiencies caused by bipartite matching with noisy pseudo-labels. 
Based on Semi-DETR, Sparse Semi-DETR~\cite{shehzadi2024sparse} improves object query quality with a query refinement module, improving the detection of small and obscured objects.
To address assignment ambiguity, ARSL~\cite{liu2023ambiguity} proposes combining Joint-Confidence Estimation and Task-Separation Assignment strategies. TMR-RD-v2~\cite{marvasti2024training} refines model weights dynamically and maintains teacher-student model distinctiveness. 

However, all the methods mentioned above are designed for the general scenes. How to unleash the potential of the semi-supervised framework in aerial scenes has yet to be fully explored.

\subsection{Oriented Object Detection}

Unlike general object detectors~\cite{girshick2015fast,ren2015faster,liu2016ssd,redmon2016you}, representing objects with Horizontal Bounding Boxes (HBBs), oriented object detectors use Oriented Bounding Boxes (OBBs) to capture the orientation of objects, which is practical for detecting aerial objects. In recent years, numerous methods~\cite{yang2021r3det,li2023large,lyu2022rtmdet,dai2022ao2,xu2020gliding,luo2023pointobb,feng2022weakly,yu2024spatial,cheng2022anchor,cheng2023towards} have been developed to enhance the performance of oriented object detection. For instance, CSL~\cite{yang2022arbitrary} addresses the out-of-bound issue by transforming the angle regression problem into a classification task. R$^3$Det~\cite{yang2021r3det} improves detection speed by predicting HBBs in the first stage and then aligning features in the second stage to detect oriented objects. Oriented R-CNN~\cite{xie2021oriented} introduces a simplified oriented region proposal network and uses midpoint offsets to represent arbitrarily oriented objects. ReDet~\cite{han2021redet} incorporates rotation-equivariant networks into the detector to extract rotation-equivariant features. Oriented RepPoints~\cite{li2022oriented} incorporates a quality assessment module and a sample assignment scheme for adaptive point learning, which helps obtain non-axis features from neighboring objects while ignoring background noise. LSKNet~\cite{li2023large} extends the large and selective kernel mechanisms to improve the performance. The recent method COBB~\cite{xiao2024theoretically} employs nine parameters derived from continuous functions based on the outer HBB and OBB area. 

The above methods typically focus on the fully supervised paradigm, requiring expensive labeling costs. Thus, several weakly supervised object detectors have been proposed recently, such as point-based methods~\cite{luo2023pointobb}, horizontal bounding boxes-based methods~\cite{yu2024h2rbox,yang2022h2rbox} or hybrid-based methods~\cite{wu2024relational}. For example, H2RBox~\cite{yu2024h2rbox} learns the rotation via self-supervised learning, whose loss measures the consistency of the predicted angles in two different views. Recently, Relational Matching~\cite{wu2024relational} introduces a weak-semi supervised setting, which comprises some fully labeled and some weakly labeled data with a single point. However, these methods are either inferior to their fully supervised counterparts or fail to leverage completely unlabeled data for performance improvement. In this paper, we explore semi-supervised oriented object detection, which reduces annotation costs and boosts detectors, even those well-trained on large-scale labeled data, by utilizing additional unlabeled data.

\subsection{Semi-Supervised Oriented Object Detection} 
Our conference work~\cite{hua2023sood} SOOD is the first to explore the semi-supervised oriented object detection task. It contains two key components: the Rotation-aware
Adaptive Weighting (RAW) loss is used to dynamically weigh the pseudo-label-prediction pairs, and the Global Consistency (GC) is designed to establish a many-to-many relationship between the sets of pseudo-labels and predictions. 

In this paper, we make the following extensions: \textbf{1)} We propose a Simple Instance-aware Dense Sampling (SIDS) strategy, which allows us to mine challenging objects that blend into the background, ensuring that both easy and hard cases are adequately represented, significantly improving the quality of pseudo-labels. \textbf{2)} We refine the RAW loss by further considering the oriented objects' aspect ratio to adaptively weigh the importance of pseudo-labels-prediction pairs, providing a clearer direction for optimization during the semi-supervised learning. \textbf{3)} We refine the GC from the single-perspective global alignment to multi-perspective global alignment, which encompasses the alignments between the original teacher and original student, the disturbed teacher and disturbed student, and the transport plans generated from the former alignments. \textbf{4)} Besides evaluating the widely-used DOTA-V1.5 dataset in the conference version, we conduct experiments on more benchmarks, such as the DOTA-V1.0 and HRSC2016 datasets. Additionally, we perform experiments on semi-supervised oriented 3D object detection on the large-scale nuScenes dataset to demonstrate the generalization of our method. 

Through the above substantial technical improvements, our SOOD++ significantly outperforms its conference version SOOD~\cite{hua2023sood} by +2.15/1.85, +1.92/1.86, and +2.33/2.28 mAP on the DOTA-V2.0/DOTA-V1.5 dataset under the 10\%, 20\%, and 30\% labeled settings, respectively. 
For more details, please refer to Sec.~\ref{sec:method} and Sec.~\ref{sec:experiments}.

\begin{figure}[t]
	\begin{center}
		\includegraphics[width=0.96\linewidth]{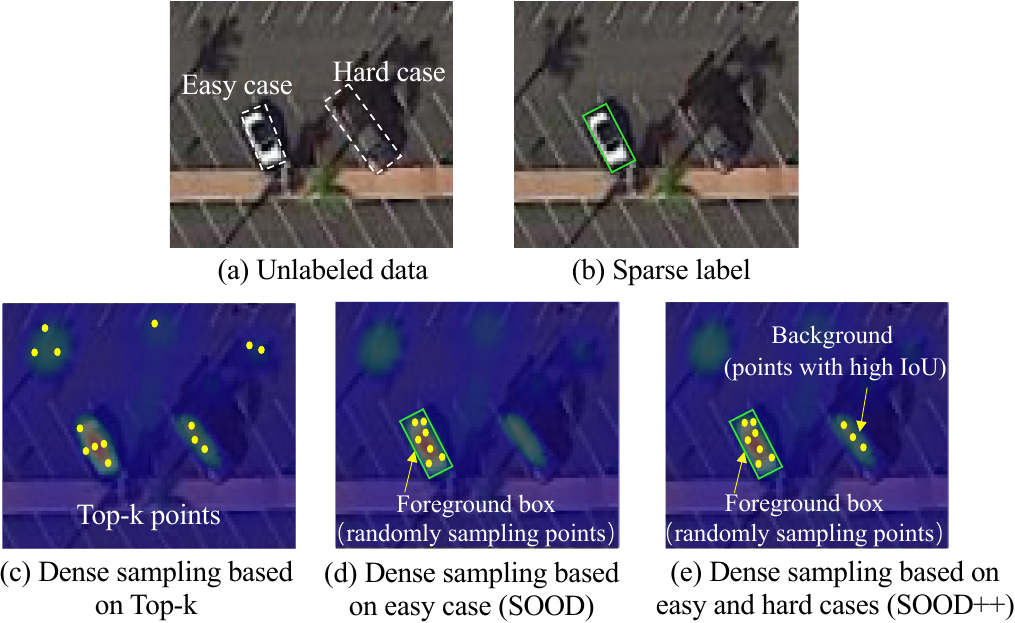}
	\end{center}
	\caption{An intuitive example illustrating different pseudo-labels: (a) Unlabeled data. (b) Sparse pseudo-label paradigm~\cite{xu2021end,liu2022unbiased}. (c) Dense sampling based on Top-K~\cite{zhou2022dense, liu2023ambiguity}. (d) Dense sampling based on foreground (easy case)~\cite{hua2023sood}. (e) The proposed dense sampling strategy incorporates both foreground and background elements. }
	\label{fig:vis_sampling}
\end{figure}

\section{Preliminary}
This section revisits the mainstream pseudo-labeling paradigm in SSOD and then introduces the concept of Monge-Kantorovich optimal transport theory~\cite{rachev1985monge}. 

\subsection{Pseudo-labeling Paradigm}
\label{sec:preliminary_pl}
Pseudo-labeling frameworks~\cite{xu2021end, liu2022unbiased, zhou2022dense,liu2023ambiguity} inherited designs from the Mean Teacher~\cite{tarvainen2017mean}, which consists of two parts, i.e., a teacher model and a student model. The teacher is an Exponential Moving Average~(EMA) of the student. They are learned iteratively by the following steps: 1) Generate pseudo-labels for the unlabeled data in a batch. The pseudo-labels are filtered from the teacher's predictions, e.g., the box coordinates and the classification scores. Meanwhile, the student makes predictions for labeled and unlabeled data in the batch. 2) Compute loss for the student's predictions. It consists of two parts: the unsupervised loss $\mathcal{L}_u$ and the supervised loss $\mathcal{L}_s$. They are computed for the unlabeled data with the pseudo-labels and the labeled data with the ground truth (GT) labels, respectively.
The overall loss $\mathcal{L}$ is the sum of them. 3) Update the parameters of the student according to the overall loss. The teacher is updated simultaneously in an EMA manner.

Based on the sparsity of pseudo-labels, pseudo-labeling frameworks can be further categorized into sparse pseudo-labeling~\cite{xu2021end, liu2022unbiased} and dense pseudo-labeling~\cite{zhou2022dense, liu2023ambiguity}, termed SPL and DPL, respectively. The SPL selects the teacher's predictions after the post-processing operations (e.g., NMS and score filtering), shown in Fig.~\ref{fig:vis_sampling}(b). It obtains sparse labels like bounding boxes and categories to supervise the student. The DPL directly samples the pixel-level post-sigmoid logits predicted by the teacher, which are dense and informative, shown in Fig.~\ref{fig:vis_sampling}(c).

\subsection{Optimal Transport}
\label{sec:preliminary_ot}

The Monge-Kantorovich Optimal Transport (OT)~\cite{rachev1985monge} aims to solve the problem of simultaneously moving items from one set to another set with minimum cost. The mathematical formulations of OT are described in detail below.

Let $\boldsymbol{X} = \{\boldsymbol{x}_i | \boldsymbol{x}_i \in \mathbb{R}^d\}_{i=1}^n$ and $\boldsymbol{Y} = \{\boldsymbol{y}_j | \boldsymbol{y}_j \in \mathbb{R}^d\}_{j=1}^n$ denote as two $d$-dimensional vector space. And we assume $\hat{\boldsymbol{x}}, \hat{\boldsymbol{y}} \in \mathbb{R}^n$ are two probability measures on $\boldsymbol{X}$ and $\boldsymbol{Y}$. The all possible transportation ways $\Gamma$ from $\boldsymbol{X}$ to $\boldsymbol{Y}$ are formed as: 
\begin{equation}
    \Gamma  = \{\boldsymbol{P}~\in\mathbb{R}^{n\times n} | \boldsymbol{P}\mathbf{1}_n = \hat{\boldsymbol{x}}, \boldsymbol{P}^T\mathbf{1}_n = \hat{\boldsymbol{y}}\},
\end{equation}
where $\mathbf{1}_n $ is a $n$-dimensional column vector with all elements equal to 1. The constraints ($\boldsymbol{P}\mathbf{1}_n = \hat{\boldsymbol{x}}$ and $\boldsymbol{P}^T\mathbf{1}_n = \hat{\boldsymbol{y}}$) ensure the probability mass from each point in the source distribution $\hat{\boldsymbol{x}}$ is accurately transported to the target distribution $\hat{\boldsymbol{y}}$, maintaining the total mass invariant. The cost between $\hat{\boldsymbol{x}}$ and $\hat{\boldsymbol{y}}$ is then defined as: 
\begin{equation}
\label{eq:wot}
    \mathcal{W}_{ot}(\hat{\boldsymbol{x}}, \hat{\boldsymbol{y}}) = \min_{\boldsymbol{P}\in \Gamma }\left \langle \boldsymbol{C}, \boldsymbol{P} \right \rangle,
\end{equation}
where $\boldsymbol{C}\in \mathbb{R}^{n \times n}$ represents the cost matrix between two sets, and $\left \langle \cdot \right \rangle$ represents inner product. In common, the OT problem is solved by the following dual formulation:
\begin{equation}
\begin{split}
     \mathcal{W}_{ot}(\hat{\boldsymbol{x}}, \hat{\boldsymbol{y}}) = \max_{\boldsymbol{\lambda},\boldsymbol{\mu}\in\mathbb{R}^n}
     \left \{
    \left \langle \boldsymbol{\lambda}, \hat{\boldsymbol{x}}  \right \rangle + 
    \left \langle \boldsymbol{\mu}, \hat{\boldsymbol{y}} \right \rangle
    \right \},
    \\
    \mathrm{s.t.}~~\boldsymbol{\lambda}_i + \boldsymbol{\mu}_j \le \boldsymbol{C}_{i, j},~\forall i, j \leq n,
\end{split}
\label{eq:dual_formulation}
\end{equation}
where $\boldsymbol{\lambda}$ and $\boldsymbol{\mu}$ are the solutions of the OT problem, which can be approximated in an iterative manner~\cite{cuturi2013sinkhorn}.

In this paper, we extend optimal transport to construct the global consistency for the teacher-student pairs, which not only enhances the robustness
of the detection process against inaccuracies in pseudo-labeling but also encourages a holistic learning approach
where both spatial information and relational context are
taken into consideration.

\section{Method}
\label{sec:method}

\begin{figure*}[t]
	\begin{center}
		\includegraphics[width=0.98\linewidth]{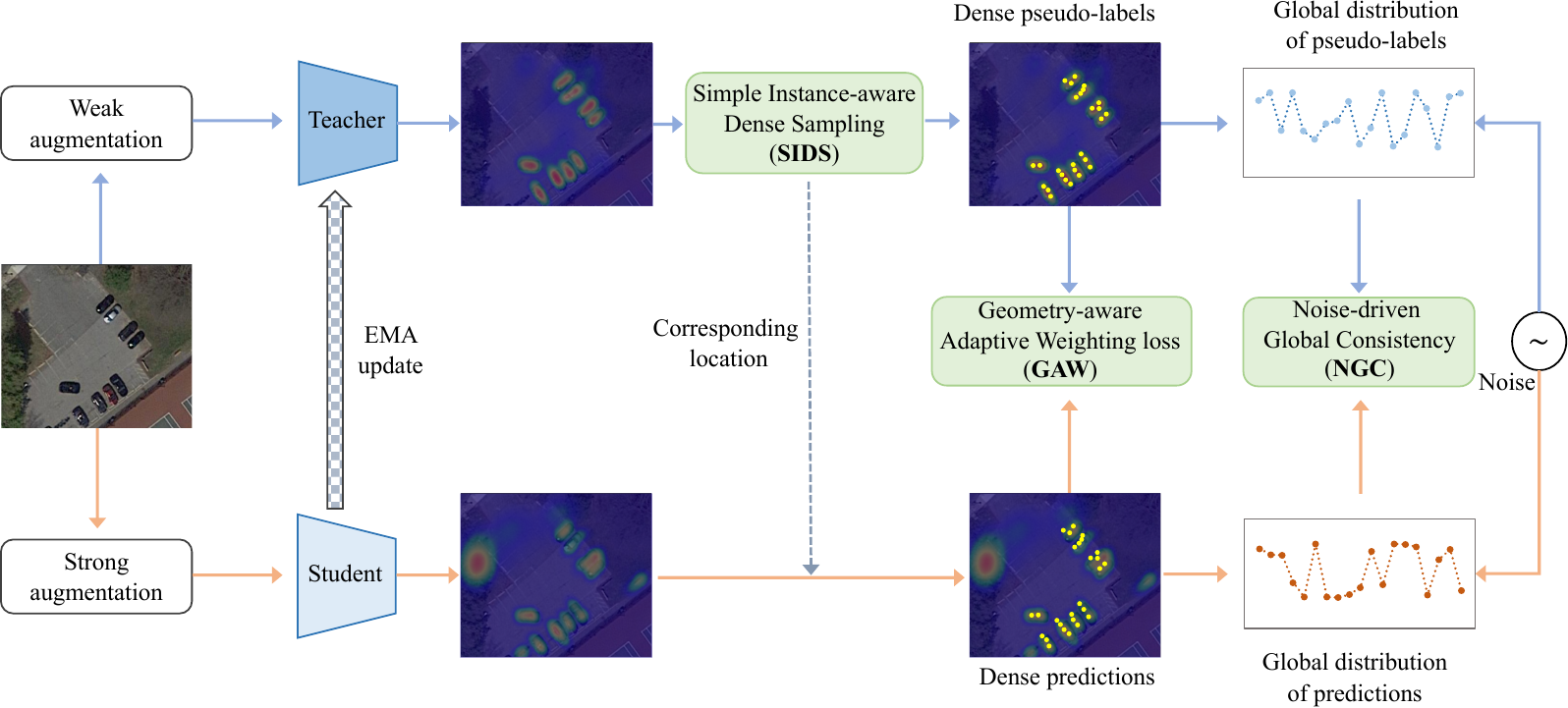}
	\end{center}
	\caption{
 The pipeline of our SOOD++. Each training batch includes labeled and unlabeled images, omitting the regular supervised part for simplicity. For the unsupervised part, we use the Simple Instance-aware Dense Sampling (SIDS) strategy to create high-quality pseudo-labels and pair them with the student's predictions. The Geometry-aware Adaptive Weighting (GAW) loss dynamically weighs each pair's unsupervised loss based on geometry information. Additionally, we treat the pseudo-labels and student predictions as global discrete distributions, measuring their similarity via Noise-driven Global Consistency (NGC). Yellow points indicate sampled dense pseudo-labels and predictions.
 }
\label{fig:pipeline}
\end{figure*}

Fig.~\ref{fig:pipeline} illustrates the pipeline of our SOOD++. Specifically, the SOOD++ is a dense pseudo-labeling framework, which consists of three key components: a Simple Instance-aware Dense Sampling (SIDS) strategy is used to construct the high-quality dense pseudo-labels, the Geometry-aware Adaptive Weighting (GAW) loss and Noise-driven Global Consistency (NGC) are adopted to measure the discrepancy between the teacher and student from an individual-level and global-level perspectives, respectively.

\subsection{The Overall Framework}
\label{sec:overall}

Aerial objects are usually dense and small, whereas the sparse-pseudo-labeling paradigm might miss massive potential objects. Thus, we choose the dense pseudo-labeling paradigm. The training process includes the supervised and unsupervised parts. For the supervised part, the student is trained regularly with labeled data. For the unsupervised part, we adopt the following steps:

\begin{itemize}
    \item First, given the output of the teacher, we utilize a Simple Instance-aware Dense Sampling (SIDS) strategy to generate the comprehensive dense pseudo-labels $\mathcal{P}^t$. We also select the predictions $\mathcal{P}^s$ at the same positions of the student. Therefore, we obtain the teacher-student pairs (i.e., $\mathcal{P}^t$ and $\mathcal{P}^s$).

    \item Next, we use the proposed Geometry-aware
Adaptive Weighting (GAW) loss to dynamically weigh each teacher-student pair's unsupervised loss by leveraging the intrinsic geometric information, including the orientation gap and aspect ratio. 

    \item Then, our proposed Noise-driven Global Consistency (NGC) will construct a \textit{relaxed} constraint by viewing the $\mathcal{P}^t$ and $\mathcal{P}^s$ as two global distributions. The random noise is used to disturb the global distributions of both the teacher and student. Following this, we achieve multi-perspective global alignments, encompassing the alignments between the original teacher and student, the perturbed teacher and perturbed student, and the transport plans generated from the former alignments.
    
\end{itemize}

We adopt the widely-used rotated version of FCOS~\cite{tian2019fcos} as the teacher and student. At each pixel-level point, rotated-FCOS will predict the confidence score, centerness, and bounding box. The basic unsupervised loss contains classification loss, regression loss, and centerness loss, corresponding to the output of FCOS. We adopt smooth $\mathcal{L}_1$ loss for regression loss, Binary Cross Entropy (BCE) loss for classification and centerness losses. Based on these, our GAW and NGC will be used to measure the consistency of the teacher and student from different perspectives.

\subsection{Simple Instance-aware Dense Sampling Strategy}
\label{sec:SIDS}

Constructing the pseudo-labels is a crucial pre-processing for the semi-supervised framework. As discussed in Sec.~\ref{sec:preliminary_pl}, since objects in aerial scenes are usually small and dense, using a dense pseudo-labeling paradigm can better identify potential objects in the unlabeled data compared with the sparse pseudo-boxes paradigm.

Previous dense pseudo-labeling methods~\cite{zhou2022dense,liu2023ambiguity} typically extract a fixed number (i.e., Top-K) of dense pseudo-labels from the predicted score map (Fig.~\ref{fig:vis_sampling}(c)).  
In the conference version (SOOD)~\cite{hua2023sood}, we propose to sample dense pseudo-labels from the foreground box areas (Fig.~\ref{fig:vis_sampling}(d)), where the foreground represents the predicted results from the teacher model after post-processing (e.g., NMS). However, it has a limitation: it often overlooks dense, small objects that blend seamlessly into the background, leading to insufficient supervision. For instance, potential high-quality positive samples might align well with the target location but are overlooked due to their low confidence in the predicted score maps.

To address the above limitation, we introduce the Simple Instance-aware Dense Sampling (SIDS) strategy, which categorizes pseudo-labels into easy (from the foreground) and hard (from the background) cases and processes them accordingly (Fig.~\ref{fig:vis_sampling}(e)). Specifically, in the foreground, we randomly sample pixel-level prediction results within the region indicated by these predicted boxes with a sample ratio $\mathcal{R}$, resulting in $\mathcal{P}^e$. In the background, we find that those potential high-quality positive samples align well with the target location despite their low confidence in the score maps. This inspires us to use the Intersection over Union (IoU) score between the predicted box and ground-truth box as a criterion for discerning pseudo-labels amidst background noise. However, given the lack of ground truth in the unlabeled dataset, explicit IoU scores are unattainable. Besides, the default rotated-FCOS output consists of class scores and bounding box coordinates (i.e., center, width, and height) for each pixel, but it does not directly capture the IoU between the predicted and ground truth bounding boxes.
Therefore, we introduce an IoU estimation branch (only one 3$\times$3 convolutional layer) trained with the labeled data that predicts the IoU of each pixel for the unlabeled data. Then, we apply threshold $\mathcal{T}_h$ to choose the pixel-level points with high IoU scores from the background, resulting in dense hard pseudo-labels $\mathcal{P}^h$. 

We merge $\mathcal{P}^h$ and $\mathcal{P}^e$ to establish the final dense pseudo-labels $\mathcal{P}^t$, which will serve as the supervisory signal for the unsupervised training process. Based on $\mathcal{P}^t$, we sample the predictions from the student at the same coordinates, defined as $\mathcal{P}^s$. After obtaining $\mathcal{P}^t$ and $\mathcal{P}^s$, the semi-supervised framework aims to make $\mathcal{P}^s$ approach $\mathcal{P}^t$, and we will describe how to construct individual-level and global-level constraints in the following sections. Due to the varying number and size of objects, our sampling number varies across different scenes, better matching the distribution of the objects. 

Admittedly, IoU estimation is indeed a widely used technique. However, our objective fundamentally differs from previous methods~\cite{feng2021tood,jiang2018acquisition,li20203d}, which primarily focus on box refinement. In contrast, our method aims to mine potential dense pseudo-labels from the background in a semi-supervised framework.

\subsection{Geometry-aware Adaptive Weighting Loss}
\label{sec:GAW}

The orientation of the oriented objects is a distinct and easily identifiable characteristic, even in cases where the objects are dense and small, as shown in Fig.~\ref{fig:intro1}(a). 
Some oriented object detection methods~\cite{qian2021learning,yang2021dense,yang2022detecting} have already employed such property in loss calculation. These works are under the assumption that the angles of the labels are reliable. Strictly forcing the prediction close to the ground truth is natural.

\begin{figure}[t]
	\begin{center}
		\includegraphics[width=0.86\linewidth]{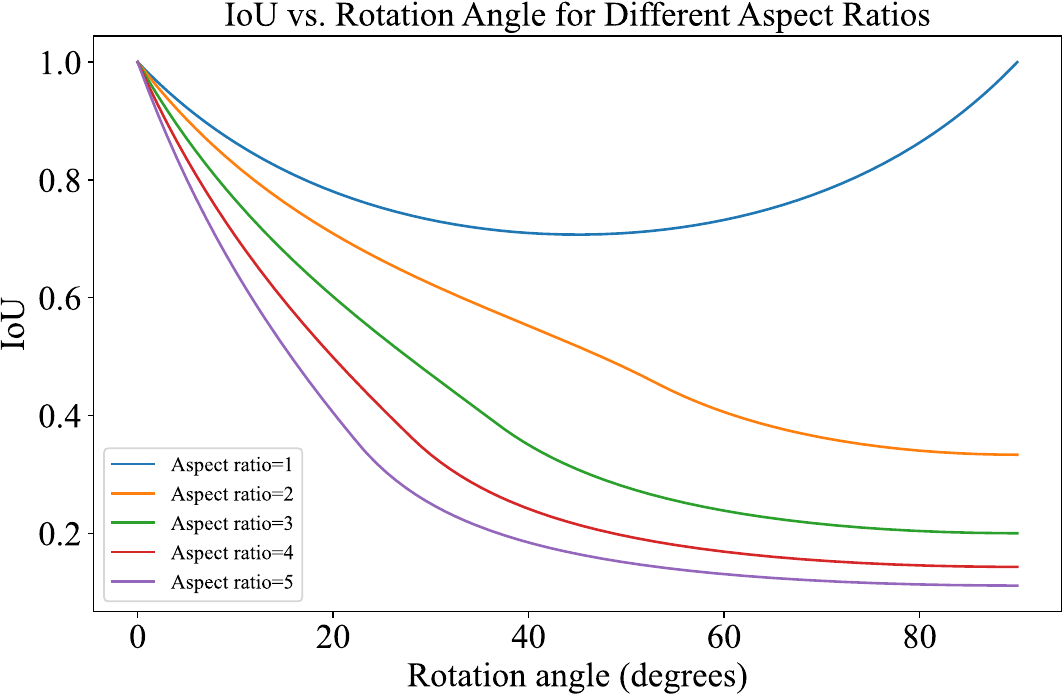}
	\end{center}
        \vspace{-10pt}
	\caption{IoU vs. Rotation angle over different aspect-ratios. }
	\label{fig:iou_ar}
\end{figure}

However, this assumption is not true under the semi-supervised setting, as the pseudo-labels may be inaccurate. Simply forcing the student to align with the teacher may lead to noise accumulation, adversely affecting the model's training. Additionally, the impact of orientation varies significantly for oriented objects with diverse aspect ratios. For instance, as illustrated in Fig.~\ref{fig:iou_ar}, an oriented object with a large aspect ratio is highly sensitive to rotation changes, and even slight misalignment can cause significant IoU change.

Based on the above observations, we propose to softly utilize orientation information and aspect ratios, enabling the semi-supervised framework to better understand the geometric priors of oriented objects. Specifically, The difference in orientation between a prediction and a pseudo-label can indicate the difficulty of the instance to some extent, which can be used to adjust the unsupervised loss dynamically. Moreover, considering that oriented objects with different aspect ratios exhibit varying sensitivities to rotation changes, it inspires us to use the aspect ratio to modulate the importance of orientation differences in indicating instance difficulty. Therefore, we propose a geometry-aware modulating factor based on these two critical geometric priors. Similar to focal loss~\cite{lin2017focal}, this factor dynamically weights the loss of each pseudo-label-prediction pair by considering its intrinsic geometric information (e.g., orientation difference and aspect ratio).

Specifically, the geometry-aware modulating factor $\omega_i^{geo}$ for the $i$-th pair is defined as:
\begin{equation}
\label{eq6}
\omega_i^{geo} = 1 + \sigma_i,
\end{equation}
\begin{equation}
\label{eq:modulating_factor}
\sigma_i = \psi \frac{\left | r_i^t - r_i^s \right | }{\pi} \frac{(a_i^t+a_i^s)}{2},  r_i^t, r _i^s \in [-\frac{\pi}{2}, \frac{\pi}{2}), a_i^t,a_i^s \geq 1,
\end{equation}
where $r_i^t$ and $r_i^s$ represent the rotation angles in radians of the $i$-th pseudo-label and prediction, respectively. $a_i^t$ and $a_i^s$ denote their aspect ratios. $\psi$ is a loss balance weight. Adding a constant to $\sigma_i$ ensures that the original unsupervised loss is preserved when the orientations of the pseudo-label and prediction coincide. The modulating factor $\sigma_i$ considers both orientation differences and average aspect ratios. A toy example is that when the difference in orientation is marginal, but the object's aspect ratio is large, the proposed factor can still efficiently highlight such a case as a possible hard example and help improve the learning process, and vice versa.

With the geometry-aware modulating factor, the overall Geometry-aware Adaptive Weighting (GAW) loss is formulated as:
\begin{equation}
\mathcal{L}_{GAW} = {\textstyle \sum_{i=1}^{N_p}} \omega_i^{geo} \mathcal{L}_u^i,
\end{equation}
where $N_p$ is the number of pseudo-labels and $\mathcal{L}_u^i$ is the basic unsupervised loss of the $i$-th pseudo-label-prediction pair.

Note that our GAW builds upon the Rotation-aware Adaptive Weighting (RAW) of the conference version~\cite{hua2023sood} by incorporating the aspect ratio as an additional factor in the weighting scheme, providing a clearer direction for optimization during the semi-supervised learning.

\subsection{Noise-driven Global Consistency}
\label{sec:NGAC}

In aerial scenes, objects often exhibit spatial relationships that are not random. For instance, vehicles might align along roads, or ships may be arranged in ports. The spatial configuration of the object set is implicitly referred to as the layout, conveying the inter-object relationships and the overarching pattern intrinsic to the image. Ideally, the layout consistency between the student's and the teacher's predictions will be preserved if each pseudo-label-prediction pair is aligned. However, such a condition is too strict and may hurt performance when there are noises in pseudo-labels. Therefore, it is reasonable to add the consistency between layouts as an additional relaxed optimization objective, encouraging the student to learn information from the teacher from a global perspective. Besides, by treating the predictions from students as a global distribution, the relations between these predictions can be regularized implicitly, which provides an implicit constraint to the student to some extent.

\begin{figure}[t]
	\begin{center}
		\includegraphics[width=0.96\linewidth]{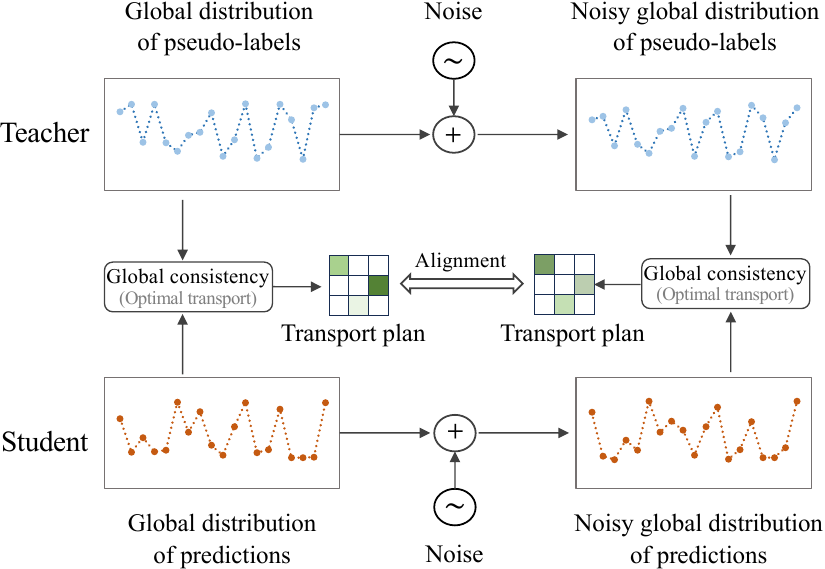}
	\end{center}
	\caption{The details of our proposed Noise-driven Global Consistency (NGC). The random noise is added to the output of the teacher and student, and then we use optimal transport to construct global consistency for the before and after disturbances. Furthermore, we make an alignment between the two transport plans. }
	\label{fig:NGC}
\end{figure}

The details of our Noise-driven Global Consistency (NGC) are shown in Fig.~\ref{fig:NGC}. Given the output of the teacher and student, we will treat them as two global distributions. First, we add different random noises to disturb both the teacher and the student distributions. Then, we propose to make multi-perspective alignments from three aspects: 1) aligning distributions between the original teacher and original student; 2) aligning distributions between the disturbed teacher and disturbed student; 3) aligning the transport plans from 1) and 2). By adding noise to both the teacher’s and student’s global distribution, we encourage the model to focus on global patterns rather than overfitting to noise or minor inconsistencies in the pseudo-labels. We argue that this regularization improves generalization, especially when the pseudo-labels are imperfect. The multi-perspective alignment ensures consistency at a different global-level perspective, maintaining the overall object structure and preventing the model from relying too heavily on potentially noisy or incorrect pseudo-labels.

Specifically, we propose to leverage the optimal transport theory to measure the global similarity of layouts between the teacher's and the student's predictions. Specifically, let $K$ denote the number of classes. For a given image\footnote{The student model samples pixel-level results at the same positions as the teacher, ensuring consistency in both the number and location.}, we define the global distributions of classification scores predicted by the teacher ($\mathbf{s}^t\in\mathbb{R}^{N_p\times K}$) and the student ($\mathbf{s}^s\in\mathbb{R}^{N_p\times K}$) as follows:
\begin{equation}
    \mathbf{d}_i^t = e^{\mathbf{s}_{i, cls(i)}^t}, \quad  \mathbf{d}_i^s = e^{\mathbf{s}_{i, cls(i)}^s},
\end{equation}
where $cls(i) \in \{0, 1, \cdots, K-1\}$ is the class index with the largest score for the $i$-th pseudo-label. And $\mathbf{s}_{i, cls(i)}^t$ indicates the score of $cls(i)$-th class for the $i$-th pseudo-label. The $\mathbf{s}_{i, cls(i)}^s$ has a similar definition but refers to the predictions of the student. The exponential used here aims for numerical stability and ensuring the value is greater than 0.

To make the model generate robust consistency, we further add noises to disturb the distributions:
\begin{equation}
    \widetilde{\mathbf{d}}^{t} = \mathbf{d}^t + \beta \boldsymbol{\epsilon}^t,  \quad     \widetilde{\mathbf{d}}^{s} = \mathbf{d}^s + \beta \boldsymbol{\epsilon}^s,
\end{equation}
where $\boldsymbol{\epsilon}^t, \boldsymbol{\epsilon}^s \sim  \mathcal{N}(0,1)$. $\beta$ is a hyper-parameter used to control the noise intensity. The original and disturbed global distributions both reflect the many-to-many relationships between the teacher and the student, highlighting the characteristics of objects in the aerial scenario. To construct the cost map for solving the OT problem, we consider the spatial distance and the score difference of each possible pair. Specifically, we measure the transport costs between pseudo-labels and predictions as follows:
\begin{equation}
    \boldsymbol{C}_{i, j} = \boldsymbol{C}_{i, j}^{dist} + \boldsymbol{C}_{i, j}^{score},
\end{equation}
\begin{equation}
    \boldsymbol{C}_{i, j}^{dist} = \frac{\|\mathbf{z}_i^t - \mathbf{z}_j^s\|_2}{\max_{1\le a, b \le N_p} \|\mathbf{z}_a^t - \mathbf{z}_b^s\|_2},
\end{equation}
\begin{equation}
    \boldsymbol{C}_{i, j}^{score} = \frac{\|\mathbf{s}_{i, cls(i)}^t - \mathbf{s}_{j, cls(j)}^s\|_1}{\max_{1 \le a, b \le N_p}
    \|\mathbf{s}_{a, cls(a)}^t - \mathbf{s}_{b, cls(b)}^s\|_1},
\end{equation}
where $\|\cdot \|_{1}$ is the $\ell^{1}$ norm of a vector and $\| \cdot \|_2$ represents the Euclidean distance.  $\mathbf{z}_i^t$ and $\mathbf{z}_j^s$ are pixel-level 2D coordinates of the $i$-th sample in the teacher and the $j$-th sample in the student, respectively.

Here, we define the global consistencies between the original (\textit{resp.} noisy) teacher and original (\textit{resp.} noisy) student as follows:
\begin{equation}
\begin{aligned}
   \mathcal{L}_{GC}(\mathbf{d}^t, \mathbf{d}^s) & = \mathcal{W}_{ot}(\mathbf{d}^{t}, \mathbf{d}^{s})\\ & = \left\langle\boldsymbol{\lambda}^{*}, 
   \frac{\mathbf{d}^t}{\|\mathbf{d}^t\|_1}\right\rangle+\left\langle\boldsymbol{\mu}^{*},
   \frac{\mathbf{d}^s}{\|\mathbf{d}^s\|_1}\right\rangle,
\end{aligned}
\end{equation}
\begin{equation}
\begin{aligned}
   \mathcal{L}_{GC}(\widetilde{\mathbf{d}}^{t}, \widetilde{\mathbf{d}}^{s}) & =     \mathcal{W}_{ot}(\widetilde{\mathbf{d}}^{t}, \widetilde{\mathbf{d}}^{s})\\ & = \left\langle\widetilde{\boldsymbol{\lambda}}^{*}, 
   \frac{\widetilde{\mathbf{d}}^{t}}{\|\widetilde{\mathbf{d}}^{t}\|_1}\right\rangle+\left\langle\widetilde{\boldsymbol{\mu}}^{*},
   \frac{\widetilde{\mathbf{d}}^{s}}{\|\widetilde{\mathbf{d}}^{s}\|_1}\right\rangle,
\end{aligned}
\end{equation} 
where ($\boldsymbol{\lambda}^{*},\boldsymbol{\mu}^{*})$ and ($\widetilde{\boldsymbol{\lambda}}^{*}, \widetilde{\boldsymbol{\mu}}^{*}$) are the solutions of Eq.~\ref{eq:dual_formulation}. We solve the OT problem by a fast Sinkhorn distance algorithm~\cite{cuturi2013sinkhorn}, obtaining the approximate solution. Based on the defined $\mathcal{L}_{GC}(\mathbf{d}^t, \mathbf{d}^s)$ and  $\mathcal{L}_{GC}(\widetilde{\mathbf{d}}^{t}, \widetilde{\mathbf{d}}^{s})$, their gradients\footnote{Note that under the semi-supervised learning framework, only the predictions of the student model have gradients.} with respect to $\mathbf{d}^s$ and $\widetilde{\mathbf{d}}^{s}$ follows:

\begin{equation}
\frac{\partial \mathcal{L}_{GC}(\mathbf{d}^t, \mathbf{d}^s)}{\partial \mathbf{d}^s}=\frac{\boldsymbol{\mu}^{*}}{\|\mathbf{d}^s\|_{1}}-\frac{\left \langle\boldsymbol{\mu}^{*}, \mathbf{d}^s\right \rangle}{\|\mathbf{d}^s\|_{1}^{2}} ,
\end{equation}
\begin{equation}
\frac{\partial \mathcal{L}_{G C}(\widetilde{\mathbf{d}}^{t}, \widetilde{\mathbf{d}}^{s})}{\partial \widetilde{\mathbf{d}}^{s}}=\frac{\widetilde{\boldsymbol{\mu}}^{*}}{\|\widetilde{\mathbf{d}}^{s}\|_{1}}-\frac{\left \langle\widetilde{\boldsymbol{\mu}}^{*}, \widetilde{\mathbf{d}}^{s}\right \rangle}{\|\widetilde{\mathbf{d}}^{s}\|_{1}^{2}} ,
\end{equation}
the gradients can be back-propagated to learn the parameters of the detector. 
In addition, there is an OT plan $\boldsymbol{P} \in \mathbb{R}^{N_p \times N_p} $ (\textit{resp.} $\widetilde{\boldsymbol{P}} \in \mathbb{R}^{N_p \times N_p} $ between the original (resp. noisy) teacher and the original (\textit{resp.} noisy) student (refer to the Eq.~\ref{eq:wot}). The OT plans reflect the mapping from the source distribution to the target one. We suggest aligning transport plans between without and with disturbance, which can evaluate the impact of noise and offer auxiliary guidance for the model (Fig.~\ref{fig:NGC}). Given the OT plan, we will multiply it with the student distribution to be better aware of the global prior and then utilize the $\mathrm{MSE}$ to measure the difference between the two types of OT plans: 
\begin{equation}
    \mathcal{L}_{plan} = \mathrm{MSE}(\boldsymbol{P} - \widetilde{\boldsymbol{P}})
\end{equation}
Overall, we obtain the training objective of NGC:
\begin{equation}
    \mathcal{L}_{NGC} = \mathcal{L}_{GC}(\mathbf{d}^t, \mathbf{d}^s) +  \mathcal{L}_{GC}(\widetilde{\mathbf{d}}^{t}, \widetilde{\mathbf{d}}^{s}) + \mathcal{L}_{plan},
\end{equation}
when we optimize such OT-based loss, the predictions converge closely to the pseudo-labels from a global perspective.

It should be noted that the feature representation is considered an ideal global distribution only when supported by a sufficient number of samples (e.g., treating a few sample points as a global distribution is unrealistic). If the sample number $N_p$ is smaller than a fixed number $\mathcal{T}_g$, we do not treat $\mathbf{s}^t$ and $\mathbf{s}^s$ as global distributions. Accordingly, the semi-supervised framework will not adopt the $\mathcal{L}_{NGC}$.

Although optimal transport theory has been explored in other methods~\cite{frogner2015learning,nguyen2021optimal,ge2021ota,yang2023hotnas}, our goal of using OT is significantly different. They mainly focus on utilizing OT to implement neural architecture search~\cite{yang2023hotnas,nguyen2021optimal}, label assignment~\cite{ge2021ota,chang2023csot}, and image matching~\cite{liu2020semantic}. However, the goal of our NGC is to establish the many-to-many relationship between the teacher and the student under the semi-supervised setting, providing robust global consistency, which is complementary to the GAW loss.

In the conference version (SOOD)~\cite{hua2023sood}, we have already introduced the concept of global consistency but only considering $\mathcal{L}_{GC}(\mathbf{d}^t, \mathbf{d}^s)$. SOOD++ further refines it by introducing random noise perturbations to both the teacher and student distributions. Then, we make multi-perspective global alignments, i.e., including $\mathcal{L}_{GC}(\mathbf{d}^t, \mathbf{d}^s)$, $\mathcal{L}_{GC}(\widetilde{\mathbf{d}}^{t}, \widetilde{\mathbf{d}}^{s})$, and $\mathcal{L}_{plan}$, as illustrated in Fig.~\ref{fig:NGC}. This noise-driven, multi-perspective consistency mechanism ensures that the model is not overly sensitive to local pseudo-label noise. Instead, it encourages learning of robust global features that reflect the overall spatial and relational structure of the scene. By aligning both the original and noisy distributions, the model effectively captures essential global patterns.

\subsection{Training Objective}

SOOD++ is trained with the proposed GAW and NGC for unlabeled data as well as the supervised loss for labeled data. The overall loss 
$\mathcal{L}$ is defined as:
\begin{equation}
    \mathcal{L} =  \underbrace{\mathcal{L}_{GAW} + \mathcal{L}_{NGC}}_{\mathcal{L}_u} + \mathcal{L}_s,
\end{equation}
where $\mathcal{L}_u$ and $\mathcal{L}_s$ indicate the semi-supervised and supervised loss, respectively. Note that the supervised loss $\mathcal{L}_s$ is the same as defined in the supervised baseline (rotated-FCOS), and our designs only modify the unsupervised part. 

\section{Experiments}

\subsection{Dataset and Evaluation Metric}

Following conventions in SSOD, for each conducted dataset, we consider two experiment settings, partially labeled data and fully labeled data, to validate the performance of a method on limited and abundant labeled data, respectively.

\begin{table*}[t]
\footnotesize
\centering
\setlength{\tabcolsep}{5.0mm}
\caption{
Comparison on the DOTA-V2.0 (val set) under 10\%, 20\%, and 30\% partially labeled data settings, using single-scale training and testing. $^\blacklozenge$/$^{\bigstar}$/$^{\bullet}$ indicate implementations using rotated Faster R-CNN/FCOS/Deformable DETR.
}
\label{tab:dota-v2.0-new-results}
\begin{tabular}{lllccc}
\toprule
Setting & Method & Publication & 10\% & 20\% & 30\% \\
\midrule
\multirow{3}{*}{Supervised-baseline}
& Rotated-Faster R-CNN~\cite{ren2015faster} & NeurIPS 15 & 39.72 & 47.58 & 49.77 \\
& Rotated-FCOS~\cite{tian2019fcos} & ICCV 19 & 38.78 & 46.74 & 49.62 \\
& Rotated Deformable DETR~\cite{zhu2020deformable} & ICLR 21 & 40.03 & 47.86 & 50.27 \\
\midrule
\multirow{8}{*}{Semi-supervised}
& Unbiased Teacher$^\blacklozenge$~\cite{liu2021unbiased} & ICLR 21 & 41.06 & 48.79 & 49.75 \\
& Soft Teacher$^\blacklozenge$~\cite{xu2021end} & ICCV 21 & 44.97 & 50.68 & 53.23 \\
& Dense Teacher$^{\bigstar}$~\cite{zhou2022dense} & ECCV 22 & 42.73 & 49.51 & 53.17 \\
& ARSL$^{\bigstar}$~\cite{liu2023ambiguity} & CVPR 23 & 43.38 & 51.13 & 53.93 \\
& TMR-RD-v2$^{\blacklozenge}$~\cite{marvasti2024training} & WACV 24 & 45.65 & 51.80 & 54.24 \\
& Sparse Semi-DETR$^{\bullet}$~\cite{shehzadi2024sparse} & CVPR 24 & 46.24 & 52.11 & 54.15 \\
\cmidrule{2-6}
& SOOD$^{\bigstar}$~\cite{hua2023sood} & CVPR 23 & 46.99 & 52.35 & 54.57 \\
& SOOD++$^{\bigstar}$(\textbf{ours}) & - & \textbf{49.14}\dplus{+2.15} & \textbf{54.27}\dplus{+1.92} & \textbf{56.90}\dplus{+2.33} \\
\bottomrule
\end{tabular}
\end{table*}

\begin{table}[t]
\scriptsize
\centering
\setlength{\tabcolsep}{4.4mm}
\caption{Comparison on the DOTA-V2.0 (val set) with single-scale training/testing under fully labeled data. Results before the arrow are the supervised baseline ($^\blacklozenge$/$^{\bigstar}$/$^{\bullet}$ for rotated Faster R-CNN/FCOS/Deformable DETR).}
\label{tab:dota-v2.0-full}
\begin{tabular}{llc}
\toprule
Method & Publication & 100\% \\
\midrule
Unbiased Teacher$^\blacklozenge$~\cite{liu2021unbiased} & ICLR 21 & 61.04 $\xrightarrow{-0.39}$ 60.65 \\
Soft Teacher$^\blacklozenge$~\cite{xu2021end} & ICCV 21 & 61.04  $\xrightarrow{+0.75}$ 61.79 \\
Dense Teacher$^{\bigstar}$~\cite{zhou2022dense} & ECCV 22 & 60.69 $\xrightarrow{+0.86}$ 61.55 \\ 
ARSL$^{\bigstar}$~\cite{liu2023ambiguity} & CVPR 23 & 60.69 $\xrightarrow{+1.37}$ 62.06 \\
TMR-RD-v2$^{\blacklozenge}$~\cite{marvasti2024training} & WACV 24 & 61.04  $\xrightarrow{+0.71}$ 61.75 \\
Sparse Semi-DETR$^{\bullet}$~\cite{shehzadi2024sparse} & CVPR 24 & 60.73 $\xrightarrow{+0.78}$ 61.51 \\
\midrule
SOOD$^{\bigstar}$~\cite{hua2023sood} & CVPR 23 & 60.69 $\xrightarrow{+2.06}$ 62.75 \\
SOOD++$^{\bigstar}$~(\textbf{ours}) & - & 60.69 $\xrightarrow{+3.62}$ \textbf{64.31} \\
\bottomrule
\end{tabular}
\end{table}

\begin{table*}[t]
\footnotesize
\centering
\setlength{\tabcolsep}{5.0mm}
\caption{
Comparison on the DOTA-V1.5 (val set) under 10\%, 20\%, and 30\% partially labeled data settings, using single-scale training and testing. $^\blacklozenge$/$^{\bigstar}$/$^{\bullet}$ indicate implementations using rotated Faster R-CNN/FCOS/Deformable DETR.
}
\label{tab:dota-v1.5-new-results}
\begin{tabular}{lllccc}
\toprule
Setting & Method & Publication & 10\% & 20\% & 30\% \\
\midrule
\multirow{3}{*}{Supervised-baseline} & Rotated-Faster R-CNN~\cite{ren2015faster} & NeurIPS 15 & 43.43 & 51.32 & 53.14 \\
& Rotated-FCOS~\cite{tian2019fcos} & ICCV 19 & 42.78 & 50.11 & 54.79 \\
& Rotated Deformable DETR~\cite{zhu2020deformable} & ICLR 21 & 43.55 & 51.41 & 53.85 \\
\midrule
\multirow{8}{*}{Semi-supervised} & Unbiased Teacher$^\blacklozenge$~\cite{liu2021unbiased} & ICLR 21 & 44.51 & 52.80 & 53.33 \\
& Soft Teacher$^\blacklozenge$~\cite{xu2021end} & ICCV 21 & 48.46 & 54.89 & 57.83 \\
& Dense Teacher$^{\bigstar}$~\cite{zhou2022dense} & ECCV 22 & 46.90 & 53.93 & 57.86 \\
& ARSL$^{\bigstar}$~\cite{liu2023ambiguity} & CVPR 23 & 47.56 & 55.05 & 58.94 \\
& TMR-RD-v2$^{\blacklozenge}$~\cite{marvasti2024training} & WACV 24 & 47.88 & 55.17 & 59.02 \\
& Sparse Semi-DETR$^{\bullet}$~\cite{shehzadi2024sparse} & CVPR 24 & 48.34 & 55.26 & 59.19 \\
\cmidrule{2-6}
& SOOD$^{\bigstar}$~\cite{hua2023sood} & CVPR 23 & 48.63 & 55.58 & 59.23 \\
& SOOD++${^\bigstar}$~(\textbf{ours}) & - & \textbf{50.48}\dplus{+1.85} & \textbf{57.44}\dplus{+1.86} & \textbf{61.51}\dplus{+2.28} \\
\bottomrule
\end{tabular}
\end{table*}

\textbf{DOTA-V1.0}~\cite{xia2018dota} is an early version of the DOTA-series datasets, which contains 15 categories identical to DOTA-V1.5, except for annotating container-crane. For the \noindent\textbf{partially labeled data setting}, we randomly sample 10\%, 20\%, and 30\% of the images from DOTA-V1.0-train as labeled data and set the remaining images as unlabeled data. For the \noindent\textbf{fully labeled data setting}, we set DOTA-V1.0-train as labeled data and DOTA-V1.0-test as unlabeled. The mAP is used as the evaluation metric.

\textbf{DOTA-V1.5}~\cite{ding2021object} DOTA-V1.5 contains more small instances~(less than 10 pixels), which makes it more challenging compared to DOTA-V1.0. For the \noindent\textbf{partially labeled data setting}, we randomly sample 10\%, 20\%, and 30\% of the images from DOTA-V1.5-train as labeled data and set the remaining images as unlabeled data. For the \noindent\textbf{fully labeled data setting}, we set DOTA-V1.5-train as labeled data and DOTA-V1.5-test as unlabeled data. The mAP is used as the evaluation metric.

\textbf{DOTA-V2.0}~\cite{ding2021object} is an enhanced version of DOTA-V1.0, incorporating larger images, additional object categories, and comprehensive annotations of tiny instances. The resulting dataset comprises 11,268 images, 1,793,658 object instances, and 18 categories in total. For the \noindent\textbf{partially labeled data setting}, we randomly sample 10\%, 20\%, and 30\% of the images from DOTA-V2.0-train as labeled data and set the remaining images as unlabeled data. For the \noindent\textbf{fully labeled data setting}, we set DOTA-V2.0-train as labeled data and DOTA-V2.0-test as unlabeled data. The mAP is used as the evaluation metric.

\textbf{HRSC2016}~\cite{liu2016ship} is a widely-used dataset for arbitrarily oriented ship detection. For the \noindent\textbf{partially labeled data setting}, we randomly sample 10\%, 20\%, and 30\% images from the training set as labeled data and set the remaining images as unlabeled data. For the \noindent\textbf{fully labeled data setting}, we set the training set as labeled data and the test set as unlabeled data. We report the AP$_{85}$ and mAP$_{50:95}$ performance on the validation set.

\textbf{DIOR-R}\cite{cheng2022anchor} is an oriented bounding box extension of the original DIOR dataset\cite{li2020object}. For the \noindent\textbf{partially labeled data setting}, we randomly sample 10\%, 20\%, and 30\% of the training images as labeled data, treating the remaining images as unlabeled.
For the \noindent\textbf{fully labeled data setting}, we use the training set as labeled data and the test set as unlabeled data. The performance is reported under the validation set.

\textbf{nuScenes}~\cite{caesar2020nuscenes} is a large-scale oriented 3D object detection dataset containing 1,000 distinct scenarios and 1.4M oriented 3D bounding boxes from 10 categories. These scenarios are recorded with a 32-beam LiDAR, six surround-view cameras, and five radars. Each scene has 20s video frames and is fully annotated with 3D bounding boxes every 0.5 seconds. For the \noindent\textbf{partially labeled data setting}, we randomly sample 1/12, 1/6, and 1/3 annotated frames from the training set as labeled data and set the remaining training frames as unlabeled data. For the \noindent\textbf{fully labeled data setting}, we set all annotated frames from the training set as labeled data and unannotated frames as unlabeled data. We use the mAP and NDS as the primary metrics. Higher values indicate better performance and the detailed information can be found in ~\cite{caesar2020nuscenes}.

\subsection{Implementation Details}

For the DOTA-V1.0, DOTA-V1.5, DOTA-V2.0, HRSC2016, and DIOR-R datasets, we take FCOS~\cite{tian2019fcos} as the representative anchor-free detector (the ResNet-50~\cite{he2016deep} with FPN~\cite{lin2017feature} used as the backbone), where BEVDet~\cite{huang2021bevdet} is adopted for the nuScenes dataset. 
Following previous works~\cite{xia2018dota,han2021align,han2021redet}, we crop the original images into 1,024$\times$1,024 patches (800$\times$800 for DIOR-R dataset). We utilize asymmetric data augmentation for unlabeled data. We use strong augmentation (random flipping, color jittering, grayscale, and Gaussian blur) for the student model and weak augmentation (random flipping) for the teacher model. The models are trained for 180k iterations on 2 $\times$ RTX3090 GPUs (except the nuScenes dataset, using 8 $\times$ V100). We set the loss balance weight $\psi$ as 50. With the SGD optimizer, the initial learning rate of 0.0025 is divided by ten at 120k and 160k. The momentum and the weight decay are set to 0.9 and 0.0001, respectively. Each GPU takes three images as input, where the proportion between unlabeled and labeled data is set to 1:2, where the sample ratio $\mathcal{R}$ and IOU threshold $\mathcal{T}_h$ are set to 0.25 and 0.1, respectively. The hyperparameter noisy intensity $\beta$ and global number threshold $\mathcal{T}_g$ are set to 0.3 and 150. These hyperparameters are kept the same across different datasets. Following previous SSOD works~\cite{zhou2022dense,liu2021unbiased}, we use the ``burn-in" strategy to initialize the teacher model.

\begin{table}[t]
\scriptsize
\centering
\setlength{\tabcolsep}{4.4mm}
\caption{Comparison on the DOTA-V1.5 (val set) with single-scale training/testing under fully labeled data. Results before the arrow are the supervised baseline ($^\blacklozenge$/$^{\bigstar}$/$^{\bullet}$ for rotated Faster R-CNN/FCOS/Deformable DETR).}
        \label{tab:dota-v1.5-full}
        \begin{tabular}{llc}
        \toprule
        Method & Publication & 100\% \\
        \midrule
        Unbiased Teacher$^\blacklozenge$~\cite{liu2021unbiased} & ICLR 21 & 66.12 $\xrightarrow{-1.27}$ 64.85 \\
        Soft Teacher$^\blacklozenge$~\cite{xu2021end} & ICCV 21 & 66.12 $\xrightarrow{+0.28}$ 66.40 \\
        Dense Teacher$^{\bigstar}$~\cite{zhou2022dense} & ECCV 22 & 65.46 $\xrightarrow{+0.92}$ 66.38 \\ 
        ARSL$^{\bigstar}$~\cite{liu2023ambiguity} & CVPR 23 & 65.46 $\xrightarrow{+1.22}$ 66.68 \\
        TMR-RD-v2$^{\blacklozenge}$~\cite{marvasti2024training} & WACV 24 & 66.12 $\xrightarrow{+0.81}$ 66.93 \\
        Sparse Semi-DETR$^{\bullet}$~\cite{shehzadi2024sparse} & CVPR 24 & 66.20 $\xrightarrow{+0.87}$ 67.07 \\
        \midrule
        SOOD$^{\bigstar}$~\cite{hua2023sood} & CVPR 23 & 65.46 $\xrightarrow{+2.24}$ 67.70 \\
        SOOD++$^{\bigstar}$~(\textbf{ours}) & - & 65.46 $\xrightarrow{+3.58}$ \textbf{69.04} \\
         \bottomrule
        \end{tabular}
            \vspace{-5pt}
\end{table}

\subsection{Main Results}
\label{sec:experiments}
We compare our method with the state-of-the-art SSOD methods~\cite{zhou2022dense,xu2021end,liu2021unbiased,liu2023ambiguity} on several challenging datasets. For a fair comparison, we re-implement these methods and have carefully tuned their hyper-parameters to obtain their best performance on oriented object detectors with the same augmentation setting.

\subsubsection{Results on the DOTA-V2.0 dataset}
We first evaluate on the large-scale DOTA-V2.0~\cite{ding2021object} dataset from the partially labeled setting and full label setting.

\vspace{0.5ex}\noindent\textbf{Partially labeled data.}
We validate SOOD++ under three partially labeled settings (10\%, 20\%, and 30\% proportion). As illustrated in Tab.~\ref{tab:dota-v2.0-new-results}, several critical observations can be highlighted:
\textbf{1)} SOOD++ consistently achieves superior performance, reaching 49.14, 54.27, and 56.90 mAP at the 10\%, 20\%, and 30\% labeling ratios, respectively, substantially exceeding our supervised baseline (rotated-FCOS\cite{tian2019fcos}) by large margins of 10.36, 7.53, and 7.28 mAP. These improvements clearly indicate that SOOD++ effectively utilizes unlabeled data, significantly boosting detection performance.
\textbf{2)} Compared to recent advanced semi-supervised approaches, including ARSL~\cite{liu2023ambiguity}, TMR-RD-v2~\cite{marvasti2024training}, and Sparse Semi-DETR~\cite{shehzadi2024sparse}, SOOD++ achieves notable superiority across all labeling settings. Particularly, under the scenario of 10\% labeled data, our method surpasses these advanced competitors by substantial margins of up to 2.90 mAP, demonstrating its remarkable ability to leverage limited labeled data in conjunction with abundant unlabeled data.
\textbf{3)} Compared with our previous conference version, SOOD~\cite{hua2023sood}, SOOD++ achieves consistent improvements of 2.15, 1.92, and 2.33 mAP under the 10\%, 20\%, and 30\% labeling ratios, respectively, reflecting substantial technical enhancements introduced in this extended version. 

\vspace{0.5ex}\noindent\textbf{Fully labeled data.}
The fully labeled data setting presents a more challenging scenario for semi-supervised learning. In this case, the detector is already trained on abundant annotated samples, and performance tends to saturate. Consequently, leveraging additional unlabeled data to yield further improvements becomes significantly more difficult, requiring both high-quality pseudo-labels and well-designed optimization strategies. As shown in Tab.~\ref{tab:dota-v2.0-full}, SOOD++ surpasses all state-of-the-art semi-supervised object detection approaches\cite{zhou2022dense,xu2021end,marvasti2024training,shehzadi2024sparse}, achieving a remarkable mAP of 64.31 and outperforming the best competing method by at least 2.25 mAP. Moreover, SOOD++ even exceeds the strong fully supervised baseline by 3.62 mAP, clearly validating the effectiveness of our semi-supervised learning framework. Interestingly, we observe that Unbiased Teacher~\cite{liu2021unbiased} experiences performance degradation when unlabeled data is introduced. We conjecture that this may be attributed to its lack of unsupervised bounding box regression loss, which is particularly critical for the precise localization required in oriented object detection.

\subsubsection{Results on the DOTA-V1.5 dataset}

We then evaluate our method on another large-scale, challenging oriented object detection dataset (DOTA-V1.5~\cite{ding2021object}).  

\vspace{0.5ex}\noindent\textbf{Partially labeled data.} Tab.~\ref{tab:dota-v1.5-new-results} presents the results of various partially labeled data settings on the DOTA-V1.5 dataset. Specifically, the proposed SOOD++ achieves state-of-the-art performance under all proportions, achieving 50.48, 57.44, and 61.51 mAP on 10\%, 20\%, and 30\% proportions, respectively, significantly surpassing our supervised baseline rotated-FCOS~\cite{tian2019fcos} by 7.70, 7.33, and 6.72 mAP. Compared with the state-of-the-art anchor-free method ARSL~\cite{liu2023ambiguity} and our conference version~\cite{hua2023sood},  we significantly outperform them by 2.92/1.85, 2.39/1.86, and 2.57/2.28 mAP, highly demonstrating the substantial improvements we made over the conference version. Compared with TMR-RD-v2~\cite{marvasti2024training} and Sparse Semi-DETR~\cite{shehzadi2024sparse}, on 10\% and 20\% proportions, our SOOD++ achieves higher performance even though our supervised baseline (rotated-FCOS) is weaker than they used (rotated-Faster R-CNN and rotated Deformable DETR). Under 30\% data proportion, SOOD++ surpasses TMR-RD-v2~\cite{marvasti2024training} and Sparse Semi-DETR~\cite{shehzadi2024sparse} by a large margin, i.e.,  2.49 and 2.32 mAP, respectively.

\vspace{0.5ex}\noindent\textbf{Fully labeled data.} As shown in Tab.~\ref{tab:dota-v1.5-full}, the results indicate that we surpass previous SSOD methods~\cite{zhou2022dense,xu2021end,marvasti2024training,shehzadi2024sparse} by at least 1.97 mAP. Besides, compared with our baseline, even though it has been trained well on massive labeled data, we still achieve 3.58 mAP improvement, demonstrating our method's ability to learn from unlabeled data. 

\begin{figure*}[t]
        \scriptsize

       \resizebox{1.0\linewidth}{!}{
        \begin{minipage}{0.5\textwidth}
         \centering
        \captionsetup{type=table} 
        \setlength{\tabcolsep}{0.8mm}
        \caption{
Comparison on the DOTA-V1.0 (val set) under 10\%, 20\%, and 30\% partially labeled data settings, using single-scale training and testing. $^\blacklozenge$/$^{\bigstar}$/$^{\bullet}$ indicate implementations using rotated Faster R-CNN/FCOS/Deformable DETR.}
        \label{tab:dota-v1.0-partial}
        \begin{tabular}{lllccc}
        \toprule
        Setting&Method &Publication&10\% &20\%&30\% \\
        \midrule
        {\multirow{3}{*}{Supervised-baseline}} &Rotated-Faster R-CNN~\cite{ren2015faster}& NeurIPS 15& 45.98 & 54.37 & 57.81\\
        &Rotated-FCOS~\cite{tian2019fcos}&ICCV 19 & 46.20 & 53.88 & 57.30 \\
        &Rotated Defor. DETR~\cite{zhu2020deformable}&ICLR 21&46.51&54.01 &57.39 \\
        \midrule
        {\multirow{6.3}{*}{Semi-supervised}}&Dense Teacher$^{\bigstar}$~\cite{zhou2022dense} & ECCV 22 & 50.87 & 58.02 & 60.91 \\
        &ARSL$^{\bigstar}$~\cite{liu2023ambiguity} & CVPR 23 & 52.89 & 59.84  & 62.92  \\
        &TMR-RD-v2$^{\blacklozenge}$~\cite{marvasti2024training} & WACV 24 & 52.81 & 59.90 & 63.12 \\
        &Sparse Semi-DETR$^{\bullet}$~\cite{shehzadi2024sparse} & CVPR 24 & 53.11 & 60.01 & 63.24\\
        \cmidrule{2-6}
        &SOOD++$^{\bigstar}$~(\textbf{ours}) & - & \textbf{54.17} & \textbf{60.53} & \textbf{64.93} \\
        
        \bottomrule
        \end{tabular}
    \end{minipage}

    \hspace{10pt}
    \begin{minipage}{0.47\textwidth}
        \centering
        \captionsetup{type=table} 
        \renewcommand\arraystretch{1.67}
        \setlength{\tabcolsep}{4.2mm}
        \caption{
Comparison on the DOTA-V1.0 (val set) with single-scale training/testing under fully labeled data. Results before the arrow are the supervised baseline ($^\blacklozenge$/$^{\bigstar}$/$^{\bullet}$ for rotated Faster R-CNN/FCOS/Deformable DETR).}
        \label{tab:dota-v1.0-full}
        \begin{tabular}{llc}
        \toprule
        Method & Publication &100\%\\
        \midrule
        Dense Teacher$^{\bigstar}$~\cite{zhou2022dense}& ECCV 22& 70.22 $\xrightarrow{+0.98}$ 71.20 \\ 
        ARSL$^{\bigstar}$~\cite{liu2023ambiguity}& CVPR 23& 70.22 $\xrightarrow{+1.41}$ 71.63 \\
        TMR-RD-v2$^{\blacklozenge}$~\cite{marvasti2024training} & WACV 24 & 70.08 $\xrightarrow{+1.60}$ 71.68 \\
        Sparse Semi-DETR$^{\bullet}$~\cite{shehzadi2024sparse} & CVPR 24& 70.29 $\xrightarrow{+1.56}$ 71.85\\
        \midrule
        SOOD++$^{\bigstar}$~(\textbf{ours})&- & 70.22 $\xrightarrow{+2.18}$ \textbf{72.40}\\
         \bottomrule
        \end{tabular}
    \end{minipage} 
    }
\end{figure*}

\begin{figure*}[t]
        \scriptsize
       \resizebox{1.0\linewidth}{!}{
        \begin{minipage}{0.58\textwidth}
         \centering
        \captionsetup{type=table}
        \setlength{\tabcolsep}{0.6mm}
\caption{Comparison on the HRSC2016 (val set) under 10\%, 20\%, and 30\% partially labeled data settings, using single-scale training and testing. $^\blacklozenge$/$^{\bigstar}$/$^{\bullet}$ indicate implementations using rotated Faster R-CNN/FCOS/Deformable DETR. We report the AP$_{85}$ and mAP$_{50:95}$. }
\label{tab:hrsc}
\begin{tabular}{lllcccccc}
\toprule
{\multirow{2}{*}{Setting}}&{\multirow{2}{*}{Method}} &{\multirow{2}{*}{Publication}} & \multicolumn{2}{c}{10\%}  & \multicolumn{2}{c}{20\%}& \multicolumn{2}{c}{30\%} \\
\cmidrule{4-9}
& & &AP &mAP &AP &mAP &AP &mAP \\
\midrule
{\multirow{2}{*}{Supervised-baseline}} &Rotated-Faster R-CNN~\cite{ren2015faster}& NeurIPS 15& 8.4 & 38.0 & 30.1 & 52.4 & 38.1 & 59.5\\
     &Rotated-FCOS~\cite{tian2019fcos}&ICCV 19 &7.3 & 37.5 & 30.2& 52.6 & 37.4 & 58.8  \\
      &Rotated Defor. DETR~\cite{zhu2020deformable}&ICLR 21 &8.0&37.9 &30.5 &52.8 & 37.5 & 58.8 \\
\midrule
{\multirow{5}{*}{Semi-supervised}}
&Dense Teacher$^{\bigstar}$~\cite{zhou2022dense} & ECCV 22 & 21.9 & 49.9 &32.8 & 59.6& 41.4 & 63.1\\
&ARSL$^{\bigstar}$~\cite{liu2023ambiguity} & CVPR 23 & 21.5&49.2&33.6&59.8&41.1&63.4 \\
&TMR-RD-v2$^{\blacklozenge}$~\cite{marvasti2024training} & WACV 24 & 21.4 & 50.1 & 34.4 & 60.1 & 41.9 & 63.5\\
&Sparse Semi-DETR$^{\bullet}$~\cite{shehzadi2024sparse} & CVPR 24 & 22.0 & 50.4 & 35.2& 60.4 & 42.0 & 63.8 \\
\cmidrule{2-9}
&SOOD++$^{\bigstar}$~(\textbf{ours})& - & \textbf{22.9} & \textbf{52.6} & \textbf{40.3} &\textbf{62.1} & \textbf{47.8} &\textbf{65.3 }\\

\bottomrule
\end{tabular}
    \end{minipage}
    \hspace{10pt}
    \begin{minipage}{0.39\textwidth}
        \centering
        \captionsetup{type=table} 
        \setlength{\tabcolsep}{2.8mm}
\caption{Comparison on the HRSC2016 (val set) with single-scale training/testing under fully labeled data. Results before the arrow are the supervised baseline ($^\blacklozenge$/$^{\bigstar}$/$^{\bullet}$ for rotated Faster R-CNN/FCOS/Deformable DETR).}
\label{tab:hrsc-full}
\begin{tabular}{llc}
\toprule
Method & Publication &100\%\\
\midrule
Dense Teacher$^{\bigstar}$~\cite{zhou2022dense}& ECCV 22& 49.5 $\xrightarrow{+3.2}$ 52.7\\ 
ARSL$^{\bigstar}$~\cite{liu2023ambiguity}& CVPR 23& 49.5 $\xrightarrow{+5.0}$ 54.5\\
TMR-RD-v2$^{\blacklozenge}$~\cite{marvasti2024training} & WACV 24 & 49.7 $\xrightarrow{+5.3}$ 55.0 \\
Sparse Semi-DETR$^{\bullet}$~\cite{shehzadi2024sparse} & CVPR 24& 49.6 $\xrightarrow{+5.4}$ 55.3 \\
\midrule

SOOD++$^{\bigstar}$~(\textbf{ours})&- & 49.5 $\xrightarrow{+8.9}$ \textbf{58.4}\\

 \bottomrule
\end{tabular}
    \end{minipage} 
    }
\end{figure*}

\subsubsection{Results on the DOTA-V1.0 dataset}

We then conduct the experiments on the DOTA-V1.0~\cite{xia2018dota} dataset, which has the same images as the DOTA-V1.5~\cite{ding2021object} dataset but provides coarse annotations (e.g., many tiny objects with fewer pixels are not annotated).

\textbf{Partially labeled data.} We evaluate our method under different labeled data proportions on the DOTA-V1.0 dataset, as shown in Tab.~\ref{tab:dota-v1.0-partial}. It is similar to the comparison of the DOTA-V2.0 and DOTA-V1.5 datasets, where our SOOD++ achieves state-of-the-art performance under all proportions. Specifically, it surpasses the Sparse Semi-DETR~\cite{shehzadi2024sparse} by 1.06, 0.52, and 1.69 mAP under 10\%, 20\%, and 30\% labeled data settings, respectively. These impressive results demonstrate that our method handles the coarse annotation dataset well.

\textbf{Fully labeled data.} Under this setting, we can explore the potential of the proposed semi-supervised framework even if it has been trained with extensive labeled data. As shown in Tab.~\ref{tab:dota-v1.0-full}, although the supervised baseline has already achieved high performance, our SOOD++ still achieves 2.18 mAP improvements on it. In addition, our method performs better compared with other SOTA semi-supervised methods~\cite{shehzadi2024sparse,marvasti2024training}.

\begin{figure*}[h]
	\begin{center}
        \includegraphics[width=0.86 \linewidth]{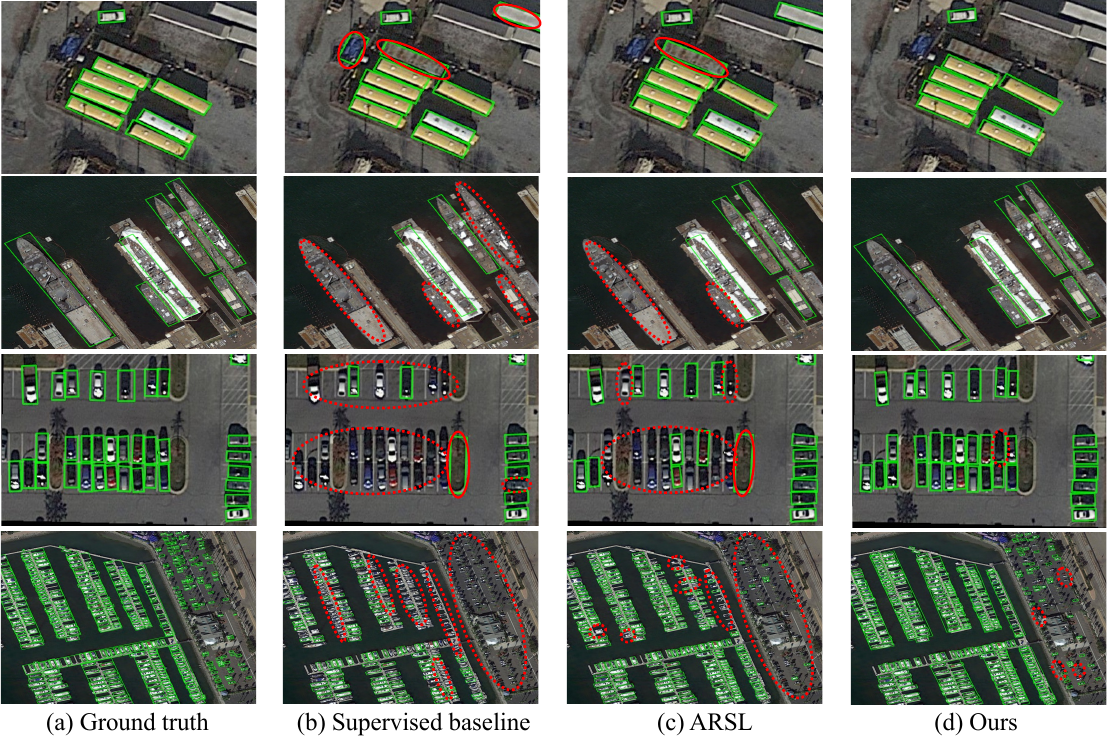}
	\end{center}
        \vspace{-10pt}
	\caption{Some typical examples from the DOTA-V1.5 dataset. We choose the dense pseudo-labeling (DPL)-based methods for visualization. 
    From left to right, there are ground truth, the supervised baseline, a SOTA DPL-based method ARSL~\cite{liu2023ambiguity}, and our proposed SOOD++. The red dashed and solid red circles represent false negative and false positive, respectively.}
	\label{fig:visualize}
\end{figure*}

\begin{figure*}[t]
\scriptsize
\resizebox{1.0\linewidth}{!}{
\begin{minipage}{0.5\textwidth}
\centering
\captionsetup{type=table}
\setlength{\tabcolsep}{0.8mm}
\caption{
Comparison on the DIOR-R (val set) under 10\%, 20\%, and 30\% partially labeled data settings, using single-scale training and testing. $^\blacklozenge$/$^{\bigstar}$/$^{\bullet}$ indicate implementations using rotated Faster R-CNN/FCOS/Deformable DETR.
}
\label{tab:dior-partial}
\begin{tabular}{lllccc}
\toprule
Setting & Method & Publication & 10\% & 20\% & 30\% \\
\midrule
\multirow{3}{*}{Supervised-baseline}
& Rotated-Faster R-CNN~\cite{ren2015faster} & NeurIPS 15 & 47.1 & 50.7 & 55.4 \\
& Rotated-FCOS~\cite{tian2019fcos} & ICCV 19 & 46.2 & 49.8 & 54.3 \\
& Rotated Deformable DETR~\cite{zhu2020deformable} & ICLR 21 & 46.3 & 50.2 & 55.7 \\
\midrule
\multirow{6}{*}{Semi-supervised}
& Dense Teacher$^{\bigstar}$~\cite{zhou2022dense} & ECCV 22 & 51.6 & 53.7 & 57.4 \\
& ARSL$^{\bigstar}$~\cite{liu2023ambiguity} & CVPR 23 & 52.0 & 55.9 & 59.2 \\
& TMR-RD-v2$^{\blacklozenge}$~\cite{marvasti2024training} & WACV 24 & 53.2 & 57.3 & 59.5 \\
& Sparse Semi-DETR$^{\bullet}$~\cite{shehzadi2024sparse} & CVPR 24 & 53.3 & 57.0 & 59.8 \\
\cmidrule{2-6}
& SOOD++$^{\bigstar}$(\textbf{ours}) & - & \textbf{56.2} & \textbf{60.1} & \textbf{63.1} \\
\bottomrule
\end{tabular}
\end{minipage}
\hspace{10pt}
\begin{minipage}{0.47\textwidth}
\centering
\captionsetup{type=table}
\renewcommand\arraystretch{1.67}
\setlength{\tabcolsep}{4.2mm}
\caption{
Comparison on the DIOR-R (val set) with single-scale training/testing under fully labeled data. Results before the arrow are the supervised baseline ($^\blacklozenge$/$^{\bigstar}$/$^{\bullet}$ for rotated Faster R-CNN/FCOS/Deformable DETR).
}
\label{tab:dior-full}
\begin{tabular}{llc}
\toprule
Method & Publication & 100\% \\
\midrule
Dense Teacher$^{\bigstar}$~\cite{zhou2022dense} & ECCV 22 & 65.7 $\xrightarrow{+1.1}$ 66.8 \\ 
ARSL$^{\bigstar}$~\cite{liu2023ambiguity} & CVPR 23 & 65.7 $\xrightarrow{+2.3}$ 68.0 \\
TMR-RD-v2$^{\blacklozenge}$~\cite{marvasti2024training} & WACV 24 & 66.1 $\xrightarrow{+2.4}$ 68.5 \\
Sparse Semi-DETR$^{\bullet}$~\cite{shehzadi2024sparse} & CVPR 24 & 65.8 $\xrightarrow{+2.5}$ 68.3 \\
\midrule
SOOD++$^{\bigstar}$~(\textbf{ours}) & - & 65.7 $\xrightarrow{+4.0}$ \textbf{69.7} \\
\bottomrule
\end{tabular}
\end{minipage}
}
    \vspace{-5pt}
\end{figure*}

\subsubsection{Results on the HRSC2016 dataset}

This part evaluates the proposed method on another representative aerial scenes dataset, i.e., HRSC2016~\cite{liu2016ship}. Notably, the commonly utilized mAP$(07)$ metric fails to adequately show the performance discrepancies among different methodologies. Consequently, we use more strict metrics to provide a more precise performance evaluation, including AP$_{85}$ and mAP$_{50:95}$. 

\vspace{0.5ex}\noindent\textbf{Partially labeled data.} As shown in Tab.~\ref{tab:hrsc}, our SOOD++ consistently achieves better performance than other semi-supervised methods~\cite{marvasti2024training,shehzadi2024sparse} across a range of settings. Especially under the strict metric (i.e., AP$_{85})$, our method shows a more pronounced improvement, which surpasses Sparse Semi-DETR~\cite{shehzadi2024sparse} by 5.1 and 5.8 mAP under 20\% and 30\% labeled data settings. 

\vspace{0.5ex}\noindent\textbf{Fully labeled data.} As shown in Tab.~\ref{tab:hrsc-full}, SOOD++ continues to demonstrate strong performance even when all annotations are available. Notably, it surpasses the supervised baseline by a substantial margin of 8.9 AP$_{85}$—far exceeding the improvements brought by prior SSOD methods, which range from only 3.2 to 5.4. This suggests that SOOD++ is not only effective in low-label regimes but also complements full supervision with more accurate and denser pseudo-labels, leading to superior final performance.

\subsubsection{Results on the DIOR-R Dataset}

We finally evaluate SOOD++ on the DIOR-R dataset~\cite{cheng2022anchor}, a large-scale benchmark with complex aerial scenes.

\vspace{0.5ex}\noindent\textbf{Partially labeled data.} 
Tab.~\ref{tab:dior-partial} shows that SOOD++ consistently outperforms prior semi-supervised methods~\cite{marvasti2024training,shehzadi2024sparse} across 10\%, 20\%, and 30\% labeled settings. Notably, our method achieves 63.1 mAP under the 30\% split, significantly surpassing Sparse Semi-DETR~\cite{shehzadi2024sparse} by 3.3 points and demonstrating superior effectiveness in low-label regimes.

\vspace{0.5ex}\noindent\textbf{Fully labeled data.} Similar to the above datasets, we also compare the performance under the fully labeled data setting. As summarized in Tab.~\ref{tab:dior-full}, 
 our method achieves 69.7 mAP, representing a 4.0-point gain over the supervised baseline. This indicates the scalability and transferability of SOOD++ across datasets and annotation densities.

\begin{table*}
    \scriptsize
    \caption{The generalization ability of our SOOD++. The results are reported with single-scale training and testing.}
    \vspace{-5pt}
       \resizebox{0.98\linewidth}{!}{

    \begin{subtable}[htbp]{0.45\linewidth} 
\scriptsize
\centering
\setlength{\tabcolsep}{2.6mm}
\renewcommand\arraystretch{1.28}
\vspace{0.3pt}
\caption{The effectiveness of SOOD++ on different methods under DOTA-V1.5 (val set) fully labeled data.}
\label{tab:ablation_detectors}
\begin{tabular}{llcc}
\toprule
Detector & Publication & Method & mAP \\
\midrule
\multirow{2}{*}{CFA~\cite{guo2021beyond}} & \multirow{2}{*}{CVPR 2021} & Supervised & 65.75  \\
 & & SOOD++~(\textbf{ours}) & \textbf{68.49} \\
 
\midrule
\multirow{2}{*}{Oriented R-CNN~\cite{xie2021oriented}}  & \multirow{2}{*}{ICCV 21} & Supervised & 67.26 \\
 & & SOOD++~(\textbf{ours}) & \textbf{69.50} \\

\bottomrule
\end{tabular}
\end{subtable}

\hspace{10pt}
\begin{subtable}[htbp]{0.5\linewidth} 
\scriptsize
\centering
\setlength{\tabcolsep}{2.8mm}
\caption{Experimental results on DOTA-V1.5 (test set). * indicates the strong supervised baseline we implemented.}
\label{tab:dota-v1.5-test}
\begin{tabular}{lccc}
\toprule
Setting & Method & Publication &mAP\\
\midrule
{\multirow{4.3}{*}{Supervised SOTA}} 
& DCFL~\cite{xu2023dynamic} & CVPR 23 & 71.03\\
& LSKNet~\cite{li2023large} & ICCV 23 & 70.26\\
& PKINet~\cite{cai2024poly} & CVPR 24 & \textbf{71.47 }\\
\cmidrule{2-4}
& Strong Oriented R-CNN* & - & 70.66\\
\midrule
Semi-supervised & SOOD++~(\textbf{ours})&- & \textbf{72.48}\\

 \bottomrule
        \end{tabular}
    \end{subtable}
}
    \vspace{-5pt}
\end{table*}

\subsection{Visualization Analysis}

We further visualize the qualitative results in Fig.~\ref{fig:visualize}. We choose the dense pseudo-labeling (DPL)-based methods for visualization. Specifically, the first sample is a sparse and relatively simpler case, and most semi-supervised methods can work well. However, when facing challenging cases, e.g., the second sample is the large aspect ratio case, and the third and fourth samples are two dense and small cases, we can find that compared with ARSL~\cite{liu2023ambiguity}, our method presents better visualization. We believe this is because SOOD++ can exploit more potential semantic information from unlabeled data by leveraging the geometry prior and global consistency. These qualitative results further highlight the effectiveness of our method.

\subsection{The Generalization Ability}

This section evaluates the generalization of our method across detectors (e.g., CFA~\cite{guo2021beyond}, Oriented R-CNN~\cite{xie2021oriented}) and a 3D related task (semi-supervised multi-view 3D oriented object detection).
\subsubsection{The generalization on different detectors}

To examine the adaptability of our method, we conduct experiments on a suite of representative oriented object detectors under the fully labeled data setting. Specifically, we choose the CFA~\cite{guo2021beyond} and Oriented R-CNN~\cite{xie2021oriented} for evaluation, as shown in Tab.~\ref{tab:ablation_detectors}. When trained with the full DOTA-V1.5 training data, the fully supervised Oriented R-CNN~\cite{xie2021oriented} achieves 67.26 mAP, which is indeed a superior performance. Even so, our method improves it by 2.24 mAP and reaches 69.50 mAP by leveraging the unlabeled data. In addition, our method also reports 2.74 mAP improvement on the CFA~\cite{guo2021beyond}. These results are significant, demonstrating our method's ability to seamlessly integrate with diverse detectors and effectively leverage the intrinsic strengths of the proposed key designs.

We further implement a strong supervised baseline (oriented R-CNN with EVA pre-trained weight~\cite{fang2023eva}) and combine it with our SOOD++ to compare with SOTA oriented detectors~\cite{li2023large,xu2023dynamic,cai2024poly}, as shown in Tab.~\ref{tab:dota-v1.5-test}. Despite the strong baseline achieving 70.66 mAP on the DOTA-V1.5 test set, our approach shows a 1.82 mAP improvement by leveraging unlabeled data\footnote{We utilize DOTA-V2.0~\cite{ding2021object} dataset as the unlabeled data in this setting.}. It outperforms the PKINet~\cite{cai2024poly} by a notable margin, pushing the new state-of-the-art.

\begin{figure}[t]
	\begin{center}
		\includegraphics[width=0.96\linewidth]{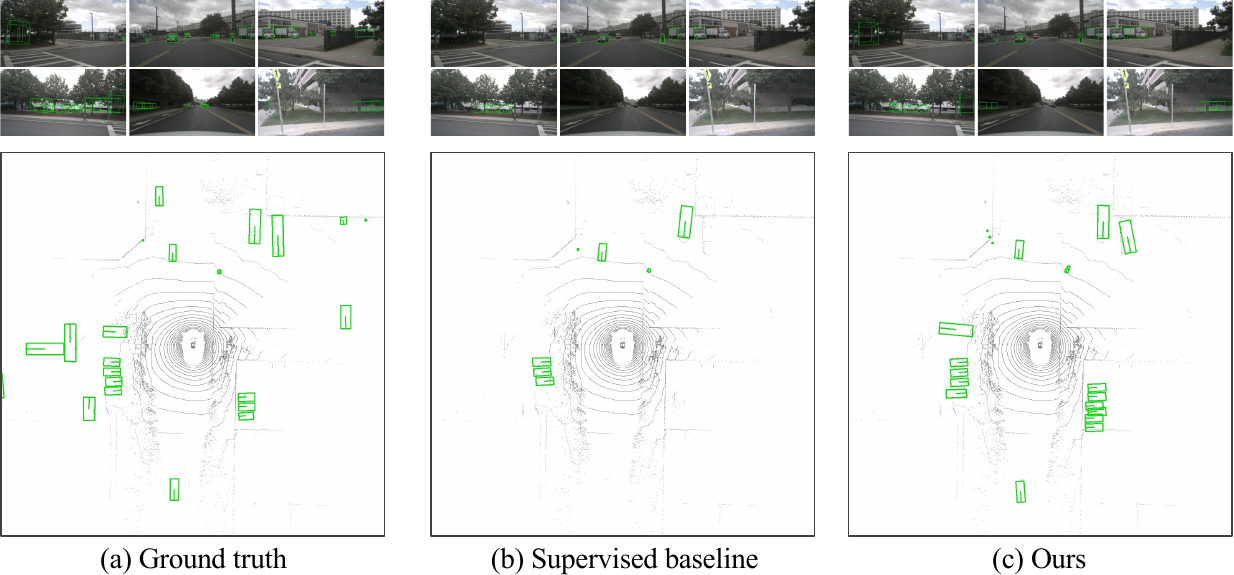}
	\end{center}
    \vspace{-10pt}
	\caption{Qualitative results for semi-supervised multi-view oriented 3D detection. Upper: Multi-view images with predicted 3D boxes. Lower: Corresponding BEV visualization.}
	\label{fig:vis_nuscenes}
        \vspace{-5pt}
\end{figure}

\subsubsection{The generalization on semi-supervised multi-view oriented 3D object detection}

We find that the multi-view oriented 3D objects are very similar to the objects from aerial scenes, as both of them have arbitrary orientations and are usually dense. In addition, the advanced multi-view oriented 3D object detection methods usually detect 3D objects by transforming the perspective from the camera view to the Bird's Eye View (BEV), similar to aerial scenes captured from a BEV. Thus, to further demonstrate the generalization of our method, we conduct experiments on the most convincing large-scale multi-view oriented 3D object detection dataset, i.e., nuScenes~\cite{caesar2020nuscenes}. As shown in Tab.~\ref{tab:nuscenes}, our method reports superior performance, outperforming the supervised baseline and ARSL~\cite{liu2023ambiguity} by a large margin in terms of mAP and NDS metrics under both partially and fully labeled data settings. We also provide detailed visualizations to present the effectiveness of our method, as presented in Fig.~\ref{fig:vis_nuscenes}. These quantitative and qualitative results prove SOOD++'s ability to be generalized to semi-supervised multi-view oriented 3D object detection.

\begin{table}[t]
\scriptsize
\centering
\setlength{\tabcolsep}{0.7mm}
\caption{The generalization on the oriented 3D object detection dataset (nuScenes~\cite{caesar2020nuscenes} val set). All semi-supervised methods use the same detector (BEVDet~\cite{huang2021bevdet}). }
\label{tab:nuscenes}
\begin{tabular}{llcccccccc}
\toprule
{\multirow{2}{*}{Setting}}&{\multirow{2}{*}{Method}} & \multicolumn{2}{c}{1/12 data}   & \multicolumn{2}{c}{1/6 data } & \multicolumn{2}{c}{1/3 data}   & \multicolumn{2}{c}{Full data} 
 \\
\cmidrule{3-10}
& &mAP  &NDS &mAP  &NDS &mAP  &NDS &mAP  &NDS  \\
\midrule
Supervised& BEVDet~\cite{huang2021bevdet} & 8.29 & 11.47 & 13.75 & 19.52 & 20.31 & 25.52 & 25.97 & 33.02\\
\midrule
{\multirow{2}{*}{Semi-}}
&ARSL~\cite{liu2023ambiguity} &8.39 & 11.90 & 15.58 & 20.91 & 21.56& 27.13 & 27.00 & 34.57 \\
&SOOD++~(\textbf{ours})  &\textbf{8.63} & \textbf{13.26} & \textbf{17.42} & \textbf{22.03} & \textbf{23.24} & \textbf{28.24} & \textbf{29.66} & \textbf{36.32}\\

\bottomrule
\end{tabular}
\end{table}

\begin{table}[t]
\scriptsize
\centering
\setlength{\tabcolsep}{2.0mm}
\caption{
Performance comparison using different pseudo-labels on the DOTA-V2.0 (val set) and DOTA-V1.5 (val set) with single-scale training and testing.
}
\label{tab:pseudo_labels}
\begin{tabular}{lccccccc}
\toprule
\multirow{2.5}{*}{Source of Pseudo-labels} & \multicolumn{3}{c}{DOTA-V2.0} & \multicolumn{3}{c}{DOTA-V1.5} \\  
\cmidrule{2-4} \cmidrule{5-7} 
& 10\% & 20\% & 30\% & 10\% & 20\% & 30\% \\
\midrule
Supervised baseline & 38.78 & 46.74 & 49.62 & 42.78 & 50.11 & 54.79 \\
+ Gemini 2.5 Flash & 39.44 & 47.09 & 50.17 & 43.62 & 50.87 & 56.03 \\
+ GPT-4.1 & 41.07 & 47.95 & 52.03 & 44.78 & 51.59 & 56.90 \\
+ SOOD++ (\textbf{ours}) & \textbf{49.14} & \textbf{54.27} & \textbf{56.90} & \textbf{50.48} & \textbf{57.44} & \textbf{61.51} \\
\bottomrule
\end{tabular}
\vspace{-5pt}
\end{table}

\subsection{Discussion on Vision-Language Models}
Recent multimodal large language models, such as GPT-4.1 and Gemini 2.5, have inspired interest in their use for pseudo-label generation in visual tasks. To assess their effectiveness for semi-supervised oriented object detection, we generate pseudo-labels using these MLLMs and compare their performance with our SOOD++ framework. As shown in Tab.~\ref{tab:pseudo_labels}, although MLLMs offer slight gains over the supervised baseline (e.g., +0.66 with Gemini 2.5 and +2.29 with GPT-4.1 under the 10\% setting on DOTA-V2.0), their performance remains substantially behind SOOD++, which achieves a remarkable improvement of +10.36 mAP. This significant gap highlights the limitations of current MLLMs in handling spatially complex and densely annotated tasks such as oriented object detection. Note that such similar findings are also observed in general vision tasks, as discussed in a recent work~\cite{ramachandran2025well}, where MLLMs struggle with fine-grained localization. These observations suggest that despite their broad visual-textual understanding, MLLMs lack the spatial precision, orientation sensitivity, and dense region awareness required by this task.
In contrast, SOOD++ incorporates task-specific strategies that are explicitly tailored for the challenges of oriented object detection. As a result, it produces substantially more reliable pseudo-labels and delivers superior semi-supervised performance across various data settings.

\begin{table}[t]
\centering
\scriptsize
\setlength{\tabcolsep}{0.8mm}
\caption{The effectiveness of Simple Instance-aware Dense Sampling (SIDS) strategy, Geometry-aware Adaptive Weighting (GAW) loss, and Noise-driven Global Consistency (NGC) on DOTA-V2.0 and DOTA-V1.5 (val set).}
\label{tab:ablation_components}
\begin{tabular}{cccccccccc}
\toprule
\multirow{2.5}{*}{Sampling strategy}  & \multirow{2.5}{*}{GAW} & \multirow{2.5}{*}{NGC} & \multicolumn{3}{c}{DOTA-V2.0} & \multicolumn{3}{c}{DOTA-V1.5} \\  
\cmidrule(r){4-6} \cmidrule(r){7-9} & & & 10\% & 20\% & 30\% & 10\% & 20\% & 30\% \\
\midrule
From Dense Teacher~\cite{zhou2022dense} & - & - & 42.73 & 49.51 & 53.17 & 46.90 & 53.93 & 57.86 \\
From SOOD~\cite{hua2023sood} & - & - & 43.15 & 50.07 & 53.54 & 47.24 & 54.07 & 57.74 \\
SIDS  & - & - & \textbf{44.32} & \textbf{51.48} & \textbf{54.45} & \textbf{48.21} & \textbf{55.54} & \textbf{58.69} \\
\midrule
{\multirow{3}{*}{SIDS}}  & \checkmark & - & 47.65 & 53.22 & 55.71 & 49.27 & 56.45 & 60.11 \\
 & - & \checkmark & 47.82 & 53.15 & 55.98 & 49.32 & 56.63 & 59.76 \\
 & \checkmark & \checkmark & \textbf{49.14} & \textbf{54.27} & \textbf{56.90} & \textbf{50.48} & \textbf{57.44} & \textbf{61.51} \\
\bottomrule
\end{tabular}
\vspace{-10pt}
\end{table}

\subsection{Ablation Study}

All the ablation experiments are performed using 10\% of labeled data with single-scale training and testing.

\subsubsection{The effectiveness of key components}

We first study the effects of the key components: the Simple Instance-aware Dense Sampling (SIDS) strategy, Geometry-aware Adaptive Weighting~(GAW) loss, and Noise-driven Global Consistency (NGC), as shown in Tab.~\ref{tab:ablation_components}: \textbf{1)} There are various dense sampling strategies for constructing dense pseudo-labels. Dense Teacher~\cite{zhou2022dense} and our conference version~\cite{hua2023sood} sample dense pseudo-labels from the entire predicted score maps and the box area of the foreground, respectively. This work not only considers the easy pseudo-labels of the foreground but also mines the potential pseudo-labels of the background, leading to the proposal of SIDS. Although our SIDS is relatively simple, compared with the strategies in~\cite{zhou2022dense,hua2023sood}, we demonstrate notable improvement, indicating the higher-quality pseudo-labels of our method. \textbf{2)} Based on SIDS, when GAW or NGC is individually incorporated, an increase in mAP is observed. This demonstrates the effectiveness of constructing the one-to-one or many-to-many (global) consistency. For example, introducing NGC significantly amplifies performance of +3.50, +1.67, and +1.53 mAP on the DOTA-V2.0 dataset, which validates its core principle of reinforcing global representation consistency through multi-perspective alignment. \textbf{3)} Combining NGC, GAW, and SIDS leads to the best overall performance, with consistent improvements across large-scale DOTA-V2.0 and DOTA-V1.5 datasets: 49.14/50.48, 54.27/57.44, and 56.90/61.51 mAP under 10\%, 20\%, and 30\% label ratios, respectively. It is reasonable, as the one-to-one and many-to-many relationships are highly complementary. These detailed ablation studies prove stable improvements in the value of each component.

Unless specified, the following ablation experiments are conducted on the large-scale DOTA-V1.5 dataset (val set).

\subsubsection{The component improvements from SOOD to SOOD++}

To intuitively compare the improvement with the conference version~\cite{hua2023sood} (SOOD), we choose SOOD as the baseline and replace each component one at a time with its corresponding new counterpart, as shown in Tab.~\ref{tab:conference_component}. 
We can observe that every new refinement module plays a crucial role in the performance boost. For example, replacing the previous sampling strategy with SIDS results in a significant improvement of +0.81 mAP at 10\%, +0.67 mAP at 20\%, and +1.13 mAP at 30\%. Similarly, replacing RAW with GAW and GC with NGC also led to notable improvements. 
When these all-new modules are combined, we achieve significant performance improvements: +1.85, +1.86, and +2.28 under 10\%, 20\%, and 30\% labeled data settings. These results clearly demonstrate the crucial contribution of each improvement in SOOD++ and how they work together to provide a more effective solution.

\begin{table}[t]
\centering
\scriptsize
\setlength{\tabcolsep}{0.5mm}
\caption{Component-wise comparison between SOOD~\cite{hua2023sood} and SOOD++ on DOTA-V1.5 (val set).}
\label{tab:conference_component}
\begin{tabular}{cccccc}
\toprule
Method&10\% & 20\% & 30\% \\
\midrule
SOOD~\cite{hua2023sood} & 48.63 & 55.58 & 59.23\\
\midrule
Replacing sampling strategy with SIDS & 49.44\dplus{+0.81} & 56.25\dplus{+0.67} & 60.36\dplus{+1.13} \\
Replacing RAW with GAW & 49.38\dplus{+0.75} & 56.29\dplus{+0.71} & 60.03\dplus{+0.80} \\
Replacing GC with NGC & 49.61\dplus{+0.98} & 56.47\dplus{+0.89} & 60.25\dplus{+1.02} \\
\midrule
SOOD++ (ours) & 50.48\dplus{+1.85} & 57.44\dplus{+1.86} & 61.51\dplus{+2.28}\\ 
\bottomrule
\end{tabular}
    \vspace{-5pt}
\end{table}

\begin{table*}
    \small
    \centering
    \caption{\dkliang{The analysis of our Simple Instance-aware Dense Sampling (SIDS) strategy. The experiments are equipped with GAW and NGC. }}
    \vspace{-10pt}
       \resizebox{0.98\linewidth}{!}{
    \begin{subtable}[htbp]{0.30\linewidth}
    \centering
    \scriptsize
    \setlength{\tabcolsep}{7.8mm} 
    \vspace{5pt}
    \caption{The effect of sample ratio $\mathcal{R}$.}
    \label{tab:ablation_sample_ratio}
    \vspace{5pt}
    \begin{tabular}{cc}
    \toprule
    Sample Ratio $\mathcal{R}$ & mAP\\
    \midrule
    0.125 & 50.13 \\
    0.25 & \textbf{50.48} \\
    0.5 & 49.92 \\
    1.0 & 49.11 \\
    
    \bottomrule
    \end{tabular}
    \end{subtable}

    \hspace*{3pt}
   \begin{subtable}[htbp]{0.31\linewidth} 
        \centering
        \scriptsize
        \setlength{\tabcolsep}{9.0mm} 
        \renewcommand\arraystretch{1.25}
        \caption{The effect of different filtering criteria adopted in the background.}
        \label{tab:criteria}
        \begin{tabular}{cc}
            \toprule
            Criteria & mAP \\
            \midrule
            Classification & 49.41 \\
            Centerness & 49.43 \\
           IoU &\textbf{50.48} \\
            \bottomrule
        \end{tabular}
    \end{subtable}

    \hspace*{3pt}
    \begin{subtable}[htbp]{0.35\linewidth} 
        \centering
        \scriptsize
        \setlength{\tabcolsep}{10.7mm} 
         \vspace{8pt}
        \caption{The effect of IoU threshold $\mathcal{T}_h$.}
        \label{tab:iou_threshold}
        \vspace{2pt}
        \begin{tabular}{cc}
            \toprule
            Threshold $\mathcal{T}_h$ & mAP \\
            \midrule
            0.05 & 50.39 \\
            0.1 &\textbf{50.48} \\
            0.2 & 50.25 \\
            0.3 & 50.17 \\
            \bottomrule
        \end{tabular}
    \end{subtable}
}

\end{table*}

\subsubsection{The analysis of Simple Instance-aware Dense Sampling (SIDS) strategy}

In this part, we discuss the design of our SIDS strategy, which is used to construct high-quality pseudo-dense labels. 

\textbf{The effect of sample ratio $\mathcal{R}$.} Tab.~\ref{tab:ablation_sample_ratio} studies the effect of the sample ratio $\mathcal{R}$ used in the foreground. The larger the number, the more pseudo-labels are sampled, and vice versa. We can see that the best performance, 50.48 mAP, is achieved when the sample ratio $\mathcal{R}$ is set to 0.25. We hypothesize that this value ensures a good balance between positive and negative pseudo-labels. Increasing it will introduce more noise pseudo-labels that harm the training process, and decreasing it will lead to the loss of some positive pseudo-labels and failure in learning. When set to 1.0, all pixel-level predictions within the instance box are sampled, which inevitably introduces a lot of noise dense pseudo-labels, leading to poor performance.

\textbf{The filtering criteria used in the background.} As discussed in Sec.~\ref{sec:SIDS}, there are many potential hard objects that merge with the background, which are hard to extract as pseudo-labels. Thus, to construct comprehensive dense pseudo-labels, we utilize the predicted IoU score as an auxiliary metric to help sample dense pseudo-labels from the background. We study the effect of using different auxiliary metrics to mine the potential labels, including classification and centerness scores of default rotated-FCOS~\cite{tian2019fcos}, as shown in Tab.~\ref{tab:criteria}. The results prove the necessity of using IoU as the filtering criterion for mining the hard cases of background. We argue that the main reason is that the IoU score directly reflects the alignment between the predicted and ground truth, making it a more comprehensive metric.

\textbf{The effect of IoU threshold $\mathcal{T}_h$. } This parameter is introduced to select the pixel-level points with high IoU scores from the background to construct the pseudo labels. Setting its value too high may under-sample, and too low may introduce too much noise. Therefore, we set $\mathcal{T}_h$ as 0.1, and it works well, as shown in Tab.~\ref{tab:iou_threshold}.

\begin{table}
    \caption{\dkliang{The analysis of our Geometry-aware Adaptive Weighting (GAW) loss. The experiments are equipped with SIDS and NGC. }}
    \vspace{-5pt}
       \resizebox{1.0\linewidth}{!}{

    \begin{subtable}[htbp]{0.53\linewidth} 
        \centering
        \scriptsize
        \setlength{\tabcolsep}{1.mm} 
        \renewcommand\arraystretch{1.2}
        \caption{The effect of used geometry information.}
        \label{tab:ablation_aspect_ratio}
        \begin{tabular}{cccc}
        \toprule
        Orientation gap & Aspect ratio & mAP\\
        \midrule
        - & - & 49.32\\
        \checkmark & - & 49.81\\
        - & \checkmark & 49.76 \\
        \checkmark & \checkmark & \textbf{50.48}\\
        \bottomrule
        \end{tabular}
    \end{subtable}

    \hspace{5pt}
      \begin{subtable}[htbp]{0.32\linewidth} 
           \scriptsize
        \centering
        \setlength{\tabcolsep}{2mm} 
        \caption{The effect of loss balance weight} $\psi$.
        \label{tab:ablation_hyper_beta}
        \begin{tabular}{cccc}
        \toprule
         $\psi$ & mAP\\
        \midrule
        w/o GAW & 49.32 \\
        1 & 49.83\\
        10 & 49.95\\
        50 & \textbf{50.48}\\
        100 & 50.13\\
        \bottomrule
        \end{tabular}
    \end{subtable}
}
  \vspace{-10pt}
\end{table}

\begin{table*}
    \centering
    \caption{The analysis of our Noise-driven Global Consistency (NGC). The experiments are equipped with SIDS strategy and GAW loss. $\mathcal{L}{GC}(\mathbf{d}^t, \mathbf{d}^s)$, $\mathcal{L}{GC}(\widetilde{\mathbf{d}}^{t}, \widetilde{\mathbf{d}}^{s})$, and $\mathcal{L}_{plan}$ denote alignments between original teacher-student, disturbed teacher-student, and the corresponding optimal transport plans, respectively.}
    \vspace{-5pt}
    \resizebox{1.\linewidth}{!}{
       
   \begin{subtable}[t]{0.49\linewidth} 
        \centering
        \setlength{\tabcolsep}{5mm} 
        \renewcommand\arraystretch{1.2}
        \caption{The effect of different alignments.}
        \label{tab:ablation_noising}
        \begin{tabular}{cc}
        \toprule
        Components  & mAP\\
        \midrule
        - &  49.27 \\
        $\mathcal{L}_{GC}(\mathbf{d}^t, \mathbf{d}^s)$& 49.83 \\
        $\mathcal{L}_{GC}(\mathbf{d}^t, \mathbf{d}^s) +  \mathcal{L}_{GC}(\widetilde{\mathbf{d}}^{t}, \widetilde{\mathbf{d}}^{s})$& 50.17 \\
        $\mathcal{L}_{GC}(\mathbf{d}^t, \mathbf{d}^s) +  \mathcal{L}_{GC}(\widetilde{\mathbf{d}}^{t}, \widetilde{\mathbf{d}}^{s}) + \mathcal{L}_{plan}$&  \textbf{50.48 }\\
        \bottomrule
        \end{tabular}
    \end{subtable}

    \begin{subtable}[t]{0.48\linewidth} 
        \centering
        \setlength{\tabcolsep}{5mm} 
        \renewcommand\arraystretch{1.2}
        \caption{The effect of different cost maps.}
        \label{tab:ablation_cost_map}
        \begin{tabular}{ccc}
        \toprule
        Distance $\boldsymbol{C}_{i, j}^{dist}$ & Score $\boldsymbol{C}_{i, j}^{score}$ & mAP\\
        \midrule
        - & - & 49.27 \\
        - & \checkmark & 49.80 \\
        \checkmark & - & 50.09 \\
        \checkmark & \checkmark & \textbf{50.48} \\
        \bottomrule
        \end{tabular}
    \end{subtable}

}
\vspace{-10pt}
\end{table*}

\begin{table}
    \small
    \centering
    \caption{Ablation study on the noisy intensity $\beta$ and global number threshold $\mathcal{T}_g$. }
    \vspace{-5pt}
    \resizebox{0.98\linewidth}{!}{
   
   \begin{subtable}[t]{0.5\linewidth} 
        \centering
        \setlength{\tabcolsep}{7mm} 
        \caption{The effect of intensity $\beta$.}
        \label{tab:ablation_beta}
        \begin{tabular}{cc}
            \toprule
            $\beta$ & mAP \\
            \midrule
            0.1 & 50.31   \\
            0.2 & 50.38  \\
            0.3 & \textbf{50.48} \\
            0.4 & 50.13  \\
            0.5 & 50.21  \\
            \bottomrule
        \end{tabular}
    \end{subtable}

    \begin{subtable}[t]{0.49\linewidth} 
        \centering
        \setlength{\tabcolsep}{7mm} 
        \caption{The effect of threshold $\mathcal{T}_g$.}
        \label{tab:ablation_tg}
        \begin{tabular}{cc}
            \toprule
            $\mathcal{T}_g$& mAP \\
            \midrule
            1 & 49.06   \\
            100 &49.83   \\
            150 & \textbf{50.48} \\
            200 & 49.95  \\
            250 & 49.68  \\
            \bottomrule
        \end{tabular}
    \end{subtable}
}
\vspace{-10pt}
\end{table}

\subsubsection{The analysis of Geometry-aware Adaptive Weighting (GAW) loss }

\textbf{The effect of geometry information in the GAW loss.} In GAW, we utilize the intrinsic geometry information (e.g., orientation gap and aspect ratios) to construct the geometry-aware modulating factor (Eq.~\ref{eq:modulating_factor}). To investigate their influence, we conduct experiments on different combinations of the GAW loss. As shown in Tab.~\ref{tab:ablation_aspect_ratio}, without the GAW, we just achieve 49.32 mAP. By individually utilizing the orientation gap or aspect ratio to construct the GAW, they report similar performance gains. When we combine the two types of information, the 50.48 mAP performance is reached. This substantial improvement confirms that using both geometric information together provides a more thorough understanding of the oriented objects' characteristics, leading to better detection performance.

\textbf{The effect of loss balance weight $\psi$.} We also study the influence of the hyper-parameter $\psi$ in GAW. As shown in Tab.~\ref{tab:ablation_hyper_beta}, we set $\psi$ to 1 and get the performance of 49.83 mAP. As $\psi$ increases, the performance of our method improves when $\psi$ varies from 1 to 50. However, further increasing it to 100 slightly hurt the performance. Therefore, we set it to 50 by default. For this observation, we conjecture that increasing the weight $\psi$ will enlarge the influence of geometry information but also amplify the impact of the teacher's inaccurate labels. Notably, GAW consistently provides performance gains regardless of the $\psi$ value.

\subsubsection{The analysis of Noise-driven Global Consistency (NGC).}

Our NGC aims to establish a \textit{relaxed} global constraint for the teacher-student pair, and this part will ablate it. 

\textbf{The effect of different global alignments.} As mentioned in Sec.~\ref{sec:NGAC}, we treat the output of the teacher and student as two global distributions and propose to use random noise to disturb the distributions. Then, we make alignment from multi-perspective, including the alignment ($\mathcal{L}_{GC}(\mathbf{d}^t, \mathbf{d}^s)$) between the original teacher and the original student, the alignment ($\mathcal{L}_{GC}(\widetilde{\mathbf{d}}^{t}, \widetilde{\mathbf{d}}^{s}) $) between the disturbed teacher and the disturbed student, and the alignment ($\mathcal{L}_{plan}$) between two OT plans generated by the two former alignments. 
Thus, we first study the effectiveness of using different alignments. As shown in Tab.~\ref{tab:ablation_noising}, only using the $\mathcal{L}_{GC}(\mathbf{d}^t, \mathbf{d}^s)$, we achieve 49.83 mAP, 0.56 mAP gain compared with the method that equipped with SIDS strategy and GAW loss, indicating the necessity of a many-to-many relationship. As expected, by further aligning the two disturbed distributions ($\mathcal{L}_{GC}(\widetilde{\mathbf{d}}^{t}, \widetilde{\mathbf{d}}^{s}) $) and two transport plans ($\mathcal{L}_{plan}$), we obtain a noticeable improvement, finally resulting in 50.48 mAP. We argue that using random noise to disturb the global distribution mitigates the detrimental effects of label noise and reinforces the model's ability to learn robust features from the unlabeled data. 

\begin{figure}[t]
	\begin{center}
		\includegraphics[width=0.96\linewidth]{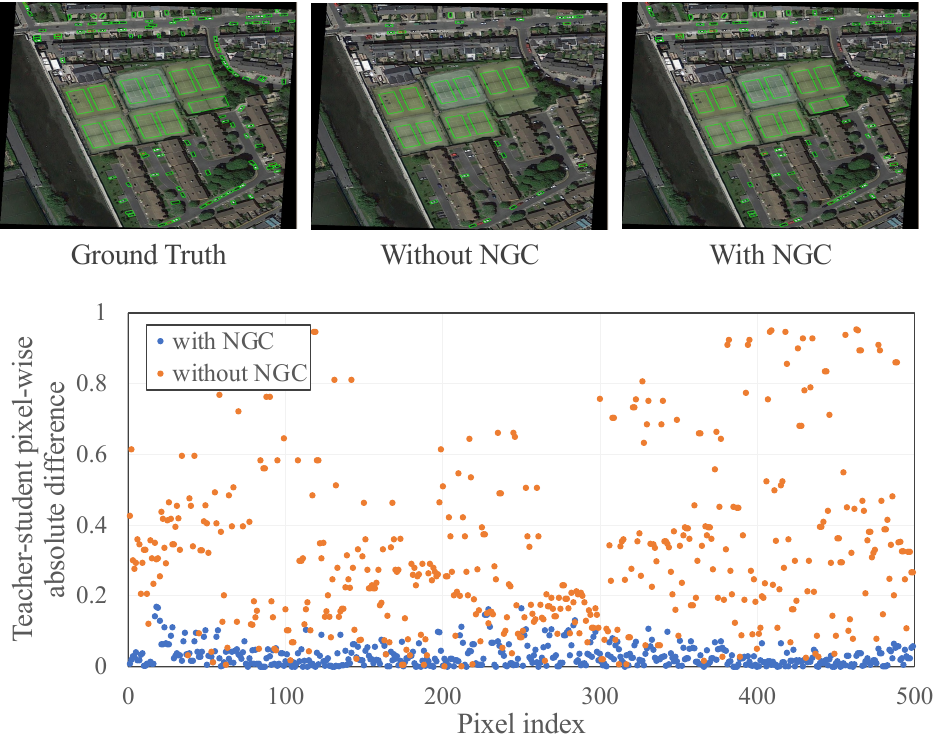}
	\end{center}
        \vspace{-10pt}
	\caption{An intuitive illustration\protect\footnotemark[4] of the distribution of the absolute difference between the teacher and student’s classification scores. Lower values indicate that the teacher and the student have more consistent distributions.}
	\label{fig:vis_NGC}
        \vspace{-5pt}
\end{figure}

\footnotetext[4]{Our method, with and without NGC, results in different numbers of pseudo-labels. Therefore, we randomly selected 500 pixel-level samples for visualization.}

\textbf{The effect of different cost maps.} To construct the cost map for solving the OT problem, we consider the spatial distance and the score difference of each possible matching pair. Thus, this part studies the effects of different cost maps of optimal transport in NGC. As shown in Tab.~\ref{tab:ablation_cost_map},
individually utilizing spatial distance or score difference results in a maximum improvement of 0.82 mAP, indicating that relying on a single type of information is insufficient for capturing the comprehensive global prior necessary for effective learning. When both spatial distance and score difference are integrated into the cost maps, the improvement is amplified to 1.21 mAP. It highlights the mutually beneficial relationship between these two factors. With their help, we effectively model the many-to-many relationship between the teacher and the student, providing informative guidance to the model.

\textbf{The effect of hyper-parameters $\beta$ and $\mathcal{T}_g$.} We conduct ablation studies to examine two hyper-parameters in our proposed NGC module: the noisy intensity $\beta$ and the global number threshold $\mathcal{T}_g$. As shown in Tab.\ref{tab:ablation_beta}, if $\beta$ is set too low (e.g., 0.1), the injected noise is insufficient to effectively encourage the model to escape local minima, limiting generalization. Conversely, excessively high values (e.g., 0.5) can disrupt global consistency, reducing performance. Therefore, we choose an intermediate value $\beta=0.3$ to strike a balance. Regarding the threshold $\mathcal{T}_g$ (Tab.\ref{tab:ablation_tg}), lower values (e.g., 100) incorrectly treat limited sampled points as global distributions, insufficiently capturing global patterns, while higher values (e.g., 250) overly restrict scenarios benefiting from global consistency. Thus, we select $\mathcal{T}_g=150$.

\textbf{Qualitative visualizations.} We further provide qualitative visualizations to analyze the effect of the proposed NGC. Specifically, we visualize detection results with and without NGC in Fig.~\ref{fig:vis_NGC}, along with the distribution of absolute difference values between the teacher and the student’s predictions. For the distribution map, lower values indicate that the teacher and the student’s distributions are more consistent. Fig.~\ref{fig:vis_NGC} clearly demonstrates that NGC can improve the consistency between the teacher and the student, leading to better prediction results.

\vspace{-5pt}
\section{Conclusion}

In this paper, we present SOOD++, a simple yet solid framework for semi-supervised oriented object detection (SSOD) tailored to aerial scenes. The integration of the sampling strategy, intrinsic geometric priors, and global consistency in SOOD++ offers a novel approach that significantly enhances detection performance by mining useful information from massive unlabeled data. Additionally, this approach provides insights that could inspire further research in both semi-supervised and oriented object detection fields. To validate the effectiveness of our method, we have conducted extensive experiments on six challenging datasets. Compared with SOTA methods, SOOD++ achieves consistent improvements under partially and fully labeled data settings.
We hope SOOD++ could be a foundation for future advancements in oriented object detection, particularly in semi-supervised learning.

\noindent\textbf{Limitation and future work.} Despite the significant progress made, several important properties of oriented objects, such as occlusions, and texture, remain largely unexplored in our work. Incorporating these factors could further enhance the model’s performance. Additionally, oriented objects and complex objects also appear in other tasks, including 3D object detection and text detection, presenting an interesting direction for future exploration.

\section*{Acknowledgments}
This work was supported by the National Natural Science Foundation of China under Grants 62225603, U2341227, and 623B2038. The authors thank Wei Xu, Xin Zhou, Xingkui Zhu, and Xingyu Jiang for their constructive comments on earlier drafts of this manuscript.

{\small
\bibliographystyle{IEEEtran}
\bibliography{egbib}
}

\end{document}